\documentclass[10pt,twocolumn]{article}
\usepackage[T1]{fontenc}
\usepackage[utf8]{inputenc}
\usepackage[letterpaper,margin=0.75in,columnsep=0.25in]{geometry}
\usepackage{microtype}
\usepackage{cite}
\usepackage{amsmath,amssymb,amsfonts}
\usepackage{algorithmic}
\usepackage{graphicx}
\usepackage{textcomp}
\usepackage{bm}
\usepackage{booktabs}
\usepackage{multirow}
\usepackage{colortbl}
\usepackage{tikz}
\usepackage{pgfplots}
\pgfplotsset{compat=1.16}
\usetikzlibrary{arrows.meta, positioning, shapes.multipart, patterns, calc, decorations.pathreplacing}
\usepgfplotslibrary{fillbetween}
\usepackage{float}
\usepackage{subcaption}
\usepackage{xcolor}
\usepackage{xurl}
\usepackage{array}
\usepackage{enumitem}
\usepackage{balance}
\usepackage[hidelinks]{hyperref}

\hypersetup{
    pdftitle={Text2Sign: A Single-GPU Diffusion Baseline for Text-to-Sign Language Video Generation},
    pdfauthor={Ruize Xia},
    pdfsubject={Text-to-sign language video generation},
    pdfkeywords={sign language generation, diffusion models, text-to-video synthesis}
}
\definecolor{ieeeblue}{rgb}{0.129,0.400,0.675}
\definecolor{tableheader}{rgb}{0.863,0.902,0.945}
\definecolor{tablerow1}{rgb}{0.961,0.973,0.988}
\definecolor{tablerow2}{rgb}{1,1,1}
\definecolor{bestcell}{rgb}{0.776,0.859,0.937}
\definecolor{accentred}{rgb}{0.698,0.094,0.169}
\definecolor{accentgreen}{rgb}{0.302,0.675,0.149}
\definecolor{accentsalmon}{rgb}{0.839,0.376,0.302}

\setlist{nosep, leftmargin=1.5em}

\newcolumntype{L}[1]{>{\raggedright\arraybackslash}p{#1}}
\newcolumntype{C}[1]{>{\centering\arraybackslash}p{#1}}
\newcolumntype{R}[1]{>{\raggedleft\arraybackslash}p{#1}}

\def\BibTeX{{\rm B\kern-.05em{\sc i\kern-.025em b}\kern-.08em
    T\kern-.1667em\lower.7ex\hbox{E}\kern-.125emX}}

\begin{document}
\title{Text2Sign: A Single-GPU Diffusion Baseline for Text-to-Sign Language Video Generation}
\author{Ruize Xia\\
\small Independent Researcher\\
\small \texttt{xiaruize0911@gmail.com}}
\date{July 2026}

\maketitle

\begin{abstract}
    
Sign language is a primary communication channel for millions of Deaf and hard-of-hearing people, yet generating signer video directly from text remains difficult because video diffusion models are expensive to train and evaluate. This paper presents \textbf{Text2Sign}, a text-conditioned diffusion architecture for short sign-language video clips, designed to operate on a single NVIDIA L4 graphics processor rather than on a multi-node training infrastructure. The model combines a frozen vision--language text encoder with a three-dimensional encoder--decoder backbone and factorized spatial and temporal attention, thereby reducing the cost of full spatio-temporal attention while preserving motion coherence. Three design choices are examined: whether transformer-style blocks improve upon convolution-only baselines, whether a frozen pretrained text encoder yields lower loss than a task-specific encoder trained from scratch under the present short-budget comparison, and whether factorized attention is competitive with full video attention. On a signer-disjoint partition of short clips extracted from How2Sign, the best short-run ablation attains a validation loss of 0.0648, while a longer-run checkpoint reaches 0.00999. A compact evaluation slice of that checkpoint yields SSIM $0.2403\pm0.0238$, PSNR $15.11\pm0.42$\,dB, and temporal consistency $1.0000\pm0.0000$; under an 8-step DDIM setting with guidance scale 5.0, the model generates a 32-frame $64\times64$ clip in 12.60\,s (2.54 frames/s) with 3.12\,GB peak inference memory on a single NVIDIA L4. In a held-out conditional denoising audit on real validation clips, removing text raises late-timestep denoising loss from 0.9875 to 0.9891, whereas shuffled prompts remain nearly indistinguishable from the intended prompt. Thus, frozen text conditioning yields a lower short-budget validation loss than the custom encoder baseline, and the revised post-revision checkpoint is qualitatively stronger than the earlier baseline in direct side-by-side inspection; however, held-out audits still show only weak prompt-specific separation. The present system remains limited to low-resolution short clips and does not yet include expert linguistic evaluation; accordingly, the reported results should be interpreted as a single-GPU research baseline rather than a complete solution to sign-language production. The code is publicly available at \url{https://github.com/xiaruize0911/text2sign}.
\end{abstract}


\section{Introduction}
\label{sec:introduction}

Accessible sign language generation constitutes a high-impact yet technically challenging problem at the intersection of computer vision and natural language processing. The World Health Organization estimates that more than 430 million people globally require rehabilitation for disabling hearing loss, with projections exceeding 700 million by 2050~\cite{who2025deafness}. In numerous settings, communication access depends on professional interpretation services---the United States Bureau of Labor Statistics reports 75,300 interpreter and translator positions in 2024, with approximately 6,900 annual openings projected over the next decade~\cite{bls2025interpreters}. These figures underscore the need for scalable sign language generation systems that can complement human interpretation by reducing friction in routine, text-mediated communication.

From a modeling perspective, automatic Sign Language Production (SLP) seeks to map spoken or written language into continuous signing sequences, whether represented as pose trajectories or video. Prior work has frequently introduced intermediate representations --- such as glosses or skeletal poses --- to stabilize long-horizon generation. Progressive Transformers~\cite{saunders2020progressive} translate text into continuous three-dimensional skeleton pose sequences, demonstrating the importance of managing temporal drift while adopting sign-specific inductive biases. In parallel, large-scale multimodal datasets such as How2Sign~\cite{duarte2021how2sign} have enabled learning from realistic sign language videos, providing both RGB frames and aligned English translations.

Diffusion models have emerged as a powerful foundation for high-fidelity conditional generation across multiple domains~\cite{ho2020ddpm, rombach2022ldm}. Video diffusion extends these ideas to temporally coherent synthesis; however, this extension incurs substantial computational and memory costs arising from spatiotemporal modeling~\cite{ho2022videodiffusion}. Transformer-based backbones for diffusion (e.g., DiT~\cite{peebles2023dit}) improve generation quality through scaling, yet na\"{i}ve attention over all video tokens remains prohibitive for practical sign language video generation.

At the same time, sign-language computing has advanced rapidly in recognition and translation. Human-keypoint-based translation systems~\cite{ko2019neural}, end-to-end recognition and translation transformers~\cite{maia2025mediapipe}, and broader surveys of text-to-sign and assistive communication systems~\cite{kahlon2023systematic,rodriguez2023assistive} provide important context for the place of generation in the pipeline. Recent recognition work has also explored depth and 3D sensing, Fourier-domain modeling, and compact dynamic feature representations~\cite{abdullahi2023idfsign,abdullahi2024fsign,abdullahi2024fourier3d,abdullahi2022prosodic,abdullahi2025redundancy}. These studies show strong progress in sign understanding. Still, they do not solve the inverse problem of producing visually plausible signing from open-ended text prompts. Direct generative studies remain relatively sparse. Recent GAN-based efforts include Indian Sign Language video synthesis from text~\cite{sreemathy2024ganvideo} and an ASL study combining GAN generators with BERT embeddings and Sora-based data augmentation~\cite{kumar2025gansbert}; however, these papers use different datasets and evaluation protocols, do not establish directly stronger results on a shared benchmark with How2Sign, and do not provide a broader directly comparable evaluation suite for sign generation. Kumar \textit{et al.} describe a preliminary quantitative evaluation without a detailed comparable metric table, while Sreemathy \textit{et al.} report a narrow SSIM/FID/MSE-style set rather than broader semantic or linguistic measures~\cite{sreemathy2024ganvideo,kumar2025gansbert}. In addition, at the time of writing, we did not identify public training code, open-source reproduction pipelines, or open-access model checkpoints for either paper. More recent work has begun to shift toward diffusion-based sign video generation. SignGen adopts latent diffusion for end-to-end signer video synthesis~\cite{qi2024signgen}, mirroring the broader generative-vision transition in which diffusion models have surpassed GANs on standard synthesis benchmarks~\cite{dhariwal2021beatgans}. Earlier text-to-sign systems were often rule-driven or avatar-oriented~\cite{filhol2016rule}, which made them easier to control but limited visual naturalism. Our work is aimed at this production setting: generating short signer video clips directly from text while staying within a single-GPU training budget.

We propose \textbf{Text2Sign}, a single-GPU text-to-sign language video diffusion baseline that directly addresses the spatio-temporal attention bottleneck. Our principal contributions are as follows:
\begin{enumerate}
    \item \textbf{Factorized spatio-temporal attention}: We decompose full 3D attention within DiT-style blocks into separate spatial and temporal components, reducing memory consumption while preserving temporal coherence for sign language motion.
    \item \textbf{Frozen pretrained text conditioning}: We employ a frozen CLIP text encoder that avoids additional text-encoder training overhead and achieves lower validation loss than a jointly trained custom encoder in the present short-budget comparison.
    \item \textbf{Comprehensive ablation study}: We systematically evaluate three architectural axes---DiT block presence, text encoder strategy, and attention factorization---quantifying their individual contributions to generation quality and computational efficiency.
    \item \textbf{Single-GPU baseline design}: The complete architecture generates temporally consistent sign language video on a single NVIDIA L4 GPU (24\,GB), demonstrating feasibility for single-GPU research workflows rather than real-time deployment.
\end{enumerate}

Our experimental results on the How2Sign dataset highlight useful architectural trade-offs for single-GPU sign-language video generation under constrained resources, while also making clear that temporally smooth motion does not yet imply reliable linguistic fidelity.

\section{Related Work}
\label{sec:related_work}

\subsection{Video Generative Methods}

Diffusion models, particularly latent variants, now form the foundation of state-of-the-art text-driven image and video synthesis. Denoising Diffusion Probabilistic Models (DDPMs) established the denoising framework~\cite{ho2020ddpm}, while Latent Diffusion Models (LDMs) demonstrated that operating in compressed latent spaces substantially reduces computational requirements while preserving visual quality~\cite{rombach2022ldm}. Dhariwal and Nichol further showed that diffusion models can surpass strong GAN baselines on standard image-synthesis benchmarks, reporting ImageNet FID 4.59 versus 6.95 for BigGAN-deep at $256\times256$ and 7.72 versus 8.43 at $512\times512$, while maintaining better distribution coverage~\cite{dhariwal2021beatgans}. The improved noise scheduling strategies introduced by Nichol and Dhariwal~\cite{nichol2021improved} further enhanced generation fidelity. Building on these foundations, LaVie~\cite{zhang2023lavie} cascades latent diffusion with temporal interpolation and super-resolution to generate high-quality video clips, and Stable Video Diffusion~\cite{blattmann2023svd} scales the approach by training in stages on large-scale datasets.

Despite their success in general scene synthesis, these models encounter significant limitations when applied to sign language generation. First, they require massive video-text corpora and considerable computational resources to capture the subtle motion patterns characteristic of signing. Second, their objective functions and sampling procedures favor broad perceptual realism over fine-grained articulatory details critical for sign language intelligibility. Third, deployment in accessibility settings --- real-time captioning, on-device inference, or resource-constrained cloud APIs --- requires architectures optimized for efficiency. Recent compression techniques such as TinyFusion-style pruning~\cite{fang2025tinyfusion} and knowledge distillation approaches~\cite{hinton2015distillation, sanh2020distilbert} indicate promising directions for reducing resource requirements, while comprehensive surveys~\cite{zhang2025videodiffusionsurvey, arxiv2025videosurvey} document the rapid expansion of this field.

The computational bottleneck in video diffusion arises primarily from attention mechanisms that scale quadratically with the number of spatiotemporal tokens. Several strategies have been proposed to mitigate this cost: factorized attention that decomposes spatial and temporal dimensions~\cite{ho2022videodiffusion}, structured pruning of diffusion transformer layers~\cite{fang2025tinyfusion}, and model compression through quantization and distillation~\cite{han2016deepcompression}. Our work builds on the factorized attention paradigm and integrates it into a DiT-enhanced 3D UNet backbone, specifically tailored to the temporal dynamics of sign language motion.

\subsection{Recent Works on Text-to-Sign}

A growing body of work addresses the conversion of written text into sign-oriented visual output, though most existing systems do not produce photorealistic signer video. VisualSign~\cite{springer2024visualsign} proposes an accessibility pipeline that processes subtitles or on-screen text to generate synchronized sign language overlays, prioritizing workflow integration for educational and broadcasting contexts by delivering via avatar-based rendering rather than full generative synthesis. Commercial platforms such as SignStream~\cite{signapse2025signstream} and Bitmovin's sign language integration~\cite{bitmovin2024signlanguage} similarly focus on low-latency APIs and avatar animation using notation systems (e.g., HamNoSys, SiGML), offering scalable overlay solutions within video players. While these systems provide substantial practical value through streaming infrastructure and synchronization capabilities, they generally circumvent the full spatiotemporal complexity inherent in modeling real human signers, trading naturalism and expressive nuance for scalability.

More direct sign-video generation studies include GAN-based pipelines. Sreemathy \textit{et al.} generate Indian Sign Language clips from text with conditional GANs~\cite{sreemathy2024ganvideo}, while Kumar \textit{et al.} explore ASL video generation using GANs, BERT text features, and Sora-augmented training data~\cite{kumar2025gansbert}. These papers are useful contextual references, but they remain difficult to benchmark directly against our setting: the sign languages, datasets, clip construction procedures, and evaluation protocols differ; neither establishes a directly stronger metric profile on a shared How2Sign benchmark; and neither provides a richer directly comparable metric suite for sign generation, since Kumar \textit{et al.} report only preliminary quantitative results while Sreemathy \textit{et al.} report a narrow SSIM/FID/MSE-style set. Moreover, at the time of writing, we did not identify public training code or open-access model checkpoints for either work, which further limits exact reproduction. More recently, SignGen uses latent diffusion for end-to-end sign video synthesis and reports significant improvements over prior state-of-the-art approaches in semantic consistency, naturalness, and expressiveness across five public sign datasets~\cite{qi2024signgen}. Taken together with the broader diffusion literature~\cite{dhariwal2021beatgans}, these results suggest that diffusion is a promising avenue for improving high-fidelity sign-video synthesis.

\section{Methodology}
\label{sec:method}

This section describes our text-conditioned diffusion framework for sign language video generation. The system accepts a text prompt $y$ and produces a sequence of RGB video frames $\mathbf{x}_0 \in \mathbb{R}^{C \times T \times H \times W}$, where $C$, $T$, $H$, and $W$ denote color channels, temporal frames, and spatial dimensions, respectively. Our design integrates a frozen CLIP-based text encoder with a 3D UNet architecture enhanced by DiT-style transformer blocks, specifically tailored for the spatio-temporal complexity of American Sign Language (ASL) as represented in the How2Sign dataset~\cite{duarte2021how2sign}.

\subsection{Dataset}
\label{subsec:dataset}

We evaluate on short clips extracted from the \textbf{How2Sign} dataset~\cite{duarte2021how2sign}, a large-scale multimodal collection of ASL videos derived from instructional content, which provides parallel RGB videos and English translations. Videos are resized to $64 \times 64$ resolution and sampled at $T = 32$ frames per clip. The processed subset is partitioned using a signer-disjoint 90/10 protocol, with the identity metadata available after preprocessing, so that no metadata-defined identity group appears in both the training and validation sets. This protocol reduces the risk of memorizing signer appearance rather than learning text-conditioned motion. The resulting split statistics are summarized in Table~\ref{tab:dataset_stats}.

\begin{table}[H]
    \centering
    \caption{Dataset Statistics After Preprocessing ($64\times64$, $T=32$) With Metadata-Based Identity-Disjoint Split}
    \label{tab:dataset_stats}
    \renewcommand{\arraystretch}{1.15}
    \begin{tabular}{@{} l r r r @{}}
        \toprule
        \rowcolor{tableheader}
        \textbf{Split} & \textbf{Clips} & \textbf{Metadata identities} & \textbf{Frames} \\
        \midrule
        \rowcolor{tablerow1}
        Train  & 3,683 & 231 & 117,856 \\
        \rowcolor{tablerow2}
        Val    & 399   & 24  & 12,768 \\
        \midrule
        \rowcolor{tablerow1}
        \textbf{Total}  & \textbf{4,082} & \textbf{255} & \textbf{130,624} \\
        \bottomrule
    \end{tabular}
\end{table}

    The processed subset contains 231 metadata-derived identity groups in the training partition and 24 in the validation partition, with zero overlap by construction. These counts refer to effective identity labels in the processed clip subset rather than to the total number of unique human signers in the original public How2Sign release. Because the data remain limited and identity representation is uneven, this protocol should be regarded as a stronger yet imperfect estimate of generalization.

\subsection{Diffusion Framework}
\label{subsec:background_diffusion}

Let $\mathbf{x}_0 \sim q(\mathbf{x}_0)$ denote a clean video sampled from the data distribution. Following the DDPM framework~\cite{ho2020ddpm}, we define a forward diffusion process that progressively corrupts $\mathbf{x}_0$ with Gaussian noise over discrete timesteps $t \in \{0, \ldots, T_\mathrm{diff}-1\}$:
\begin{equation}
    q(\mathbf{x}_t \mid \mathbf{x}_{t-1}) = \mathcal{N}\bigl(\mathbf{x}_t;\, \sqrt{1-\beta_t}\,\mathbf{x}_{t-1},\, \beta_t \mathbf{I}\bigr),
\end{equation}
where $\{\beta_t\}$ is a variance schedule. Defining $\alpha_t := 1 - \beta_t$ and $\bar{\alpha}_t := \prod_{s=0}^{t} \alpha_s$, the marginal at any timestep admits a closed-form expression:
\begin{equation}
    q(\mathbf{x}_t \mid \mathbf{x}_0) = \mathcal{N}\bigl(\mathbf{x}_t;\, \sqrt{\bar{\alpha}_t}\,\mathbf{x}_0,\, (1-\bar{\alpha}_t)\mathbf{I}\bigr),
\end{equation}
enabling direct sampling via $\mathbf{x}_t = \sqrt{\bar{\alpha}_t}\,\mathbf{x}_0 + \sqrt{1-\bar{\alpha}_t}\,\boldsymbol{\epsilon}$ with $\boldsymbol{\epsilon} \sim \mathcal{N}(\mathbf{0}, \mathbf{I})$.

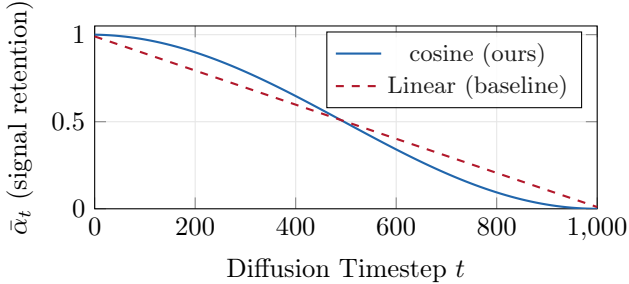
\begin{figure}[H]
    \centering
    \begin{tikzpicture}
    \begin{axis}[
        width=0.96\columnwidth,
        height=4cm,
        xlabel={Diffusion Timestep $t$},
        ylabel={$\bar{\alpha}_t$ (signal retention)},
        legend pos=north east,
        legend style={font=\small, fill=white, fill opacity=0.9},
        grid=both,
        grid style={gray!20},
        xmin=0, xmax=1000,
        ymin=0, ymax=1.05,
        every axis plot/.append style={thick},
    ]
    % Cosine schedule (cos takes degrees in pgf)
    \addplot[domain=0:1000, samples=100, color=ieeeblue, thick] 
        {(cos((x/1000 + 0.008)/(1.008) * 90))^2 / (cos(0.008/1.008 * 90))^2};
    \addlegendentry{cosine (ours)}
    % Linear schedule for comparison
    \addplot[domain=0:1000, samples=100, color=accentred, dashed, thick] 
        {max(1 - x/1000 * 0.98 - 0.01, 0.001)};
    \addlegendentry{Linear (baseline)}
    \end{axis}
    \end{tikzpicture}
    \caption{Signal retention $\bar{\alpha}_t$ under cosine (solid blue) and linear (dashed red) noise schedules. The cosine schedule preserves more low-timestep signal, helping the model learn fine details earlier.}
    \label{fig:noise_schedule}
\end{figure}

\paragraph{Noise schedule.} We adopt the cosine schedule~\cite{nichol2021improved}, which prevents premature signal destruction at low timesteps and ensures meaningful feature learning across all noise levels (Fig.~\ref{fig:noise_schedule}):
\begin{equation}
    \bar{\alpha}_t = \frac{\cos^2\!\bigl(\tfrac{t/T_\mathrm{diff}+s}{1+s} \cdot \tfrac{\pi}{2}\bigr)}{\cos^2\!\bigl(\tfrac{s}{1+s} \cdot \tfrac{\pi}{2}\bigr)},\quad \beta_t = 1 - \frac{\bar{\alpha}_t}{\bar{\alpha}_{t-1}},
\end{equation}
with offset $s$ and an upper bound on $\beta_t$.

\paragraph{Reverse process.} The generative model approximates the reverse chain $p_\theta(\mathbf{x}_{t-1} \mid \mathbf{x}_t, y)$, parameterized through a noise-prediction network $\boldsymbol{\epsilon}_\theta$:
\begin{equation}
    \boldsymbol{\mu}_\theta(\mathbf{x}_t, t, y) = \frac{1}{\sqrt{\alpha_t}} \left(\mathbf{x}_t - \frac{\beta_t}{\sqrt{1-\bar{\alpha}_t}}\,\boldsymbol{\epsilon}_\theta(\mathbf{x}_t, t, y)\right).
\end{equation}

\subsection{Text-to-Video Architecture}
\label{subsec:architecture}

\begin{figure*}[!t]
    \centering
    \resizebox{0.70\textwidth}{!}{%
    \begin{tikzpicture}[x=1cm, y=1cm,
        block/.style={draw, rounded corners=4pt, thick, align=center, minimum height=1.1cm, minimum width=3.8cm, text width=3.6cm, font=\small},
        convblock/.style={block, fill=gray!12, draw=gray!60},
        ditblock/.style={block, fill=ieeeblue!12, draw=ieeeblue!60},
        condblock/.style={block, fill=orange!10, draw=orange!60},
        ioblock/.style={block, fill=green!8, draw=green!50!black!40},
    ]
        \node[ioblock] (input) at (0,10) {\shortstack{Input Video $\mathbf{x}_{t}$\\$3\times32\times64\times64$}};
        \node[convblock] (stem) at (0,8) {\shortstack{\textbf{3D Conv Stem}\\$3\to96$ channels}};
        \node[condblock] (time) at (-5.5,1) {\shortstack{\textbf{Timestep Emb.}\\$\mathbf{e}_{t} \in \mathbb{R}^{384}$}};
        \node[condblock] (text) at (5.5,1) {\shortstack{\textbf{CLIP Text Encoder}\\(frozen) $L\times512$}};
        \node[convblock] (down1) at (0,6) {\shortstack{\textbf{Down Stage 1}\\Res3D + AdaLN\\$64\to32$, $96\text{ch}$}};
        \node[ditblock] (down2) at (0,4) {\shortstack{\textbf{Down Stage 2}\\Res3D + \textcolor{ieeeblue}{DiT Block}\\$32\to16$, $192\text{ch}$}};
        \node[ditblock] (down3) at (0,2) {\shortstack{\textbf{Down Stage 3}\\Res3D + \textcolor{ieeeblue}{DiT Block}\\$16\to8$, $384\text{ch}$}};
        \node[ditblock] (bott) at (0,0) {\shortstack{\textbf{Bottleneck}\\Res3D + \textcolor{ieeeblue}{DiT Block}\\$8\times8$, $384\text{ch}$}};
        \node[ditblock] (up3) at (0,-2) {\shortstack{\textbf{Up Stage 3}\\Skip + \textcolor{ieeeblue}{DiT Block}\\$8\to16$, $192\text{ch}$}};
        \node[ditblock] (up2) at (0,-4) {\shortstack{\textbf{Up Stage 2}\\Skip + \textcolor{ieeeblue}{DiT Block}\\$16\to32$, $96\text{ch}$}};
        \node[convblock] (up1) at (0,-6) {\shortstack{\textbf{Up Stage 1}\\Skip + Res3D\\$32\to64$, $96\text{ch}$}};
        \node[ioblock] (out) at (0,-8) {\shortstack{Output $\boldsymbol{\epsilon}_{\theta}$\\$96\to3$ channels}};
        
        % Main data flow
        \draw[-Latex, thick, black!70] (input) -- (stem);
        \draw[-Latex, thick, black!70] (stem) -- (down1);
        \draw[-Latex, thick, black!70] (down1) -- (down2);
        \draw[-Latex, thick, black!70] (down2) -- (down3);
        \draw[-Latex, thick, black!70] (down3) -- (bott);
        \draw[-Latex, thick, black!70] (bott) -- (up3);
        \draw[-Latex, thick, black!70] (up3) -- (up2);
        \draw[-Latex, thick, black!70] (up2) -- (up1);
        \draw[-Latex, thick, black!70] (up1) -- (out);
        
        % Skip connections (dashed)
        \draw[-Latex, dashed, thick, gray!60] (down3.east) -- ++(1.2,0) |- (up3.east);
        \draw[-Latex, dashed, thick, gray!60] (down2.east) -- ++(1.8,0) |- (up2.east);
        \draw[-Latex, dashed, thick, gray!60] (down1.east) -- ++(2.4,0) |- (up1.east);
        
        % Timestep conditioning (blue)
        \draw[-Latex, thick, ieeeblue, densely dashed] (time.east) -- node[above, sloped, font=\scriptsize\bfseries, text=ieeeblue] {AdaLN} (down2.west);
        \draw[-Latex, thick, ieeeblue, densely dashed] (time.east) -- node[above, sloped, font=\scriptsize\bfseries, text=ieeeblue] {AdaLN} (down3.west);
        \draw[-Latex, thick, ieeeblue, densely dashed] (time.east) -- node[above, sloped, font=\scriptsize\bfseries, text=ieeeblue] {AdaLN} (bott.west);
        \draw[-Latex, thick, ieeeblue, densely dashed] (time.east) -- node[above, sloped, font=\scriptsize\bfseries, text=ieeeblue] {AdaLN} (up3.west);
        \draw[-Latex, thick, ieeeblue, densely dashed] (time.east) -- node[above, sloped, font=\scriptsize\bfseries, text=ieeeblue] {AdaLN} (up2.west);
        
        % Text conditioning (red)
        \draw[-Latex, thick, accentred, densely dashed] (text.west) -- node[above, sloped, font=\scriptsize\bfseries, text=accentred] {Cross-Attn} (down2.east);
        \draw[-Latex, thick, accentred, densely dashed] (text.west) -- node[above, sloped, font=\scriptsize\bfseries, text=accentred] {Cross-Attn} (down3.east);
        \draw[-Latex, thick, accentred, densely dashed] (text.west) -- node[above, sloped, font=\scriptsize\bfseries, text=accentred] {Cross-Attn} (bott.east);
        \draw[-Latex, thick, accentred, densely dashed] (text.west) -- node[above, sloped, font=\scriptsize\bfseries, text=accentred] {Cross-Attn} (up3.east);
        \draw[-Latex, thick, accentred, densely dashed] (text.west) -- node[above, sloped, font=\scriptsize\bfseries, text=accentred] {Cross-Attn} (up2.east);
        
        % Legend
        \node[anchor=north west, font=\small] at (-7.5, -6.5) {
            \begin{tabular}{@{}c@{\,}l@{}}
                \tikz\draw[thick, ieeeblue, densely dashed, -Latex] (0,0) -- (0.8,0); & Timestep cond. (AdaLN) \\
                \tikz\draw[thick, accentred, densely dashed, -Latex] (0,0) -- (0.8,0); & Text cond. (Cross-Attn) \\
                \tikz\draw[dashed, thick, gray!60, -Latex] (0,0) -- (0.8,0); & Skip connections \\
            \end{tabular}
        };
    \end{tikzpicture}}
    \caption{Architecture overview of the Text2Sign 3D UNet with DiT-style transformer blocks. The network operates on 3D video tensors through an encoder--decoder structure with skip connections. DiT blocks (blue shading) apply factorized spatial--temporal attention at resolutions $16\times16$ and $8\times8$. Blue dashed arrows indicate timestep conditioning via Adaptive Layer Normalization (AdaLN); red dashed arrows indicate text conditioning via cross-attention with frozen CLIP features.}
    \label{fig:architecture}
\end{figure*}
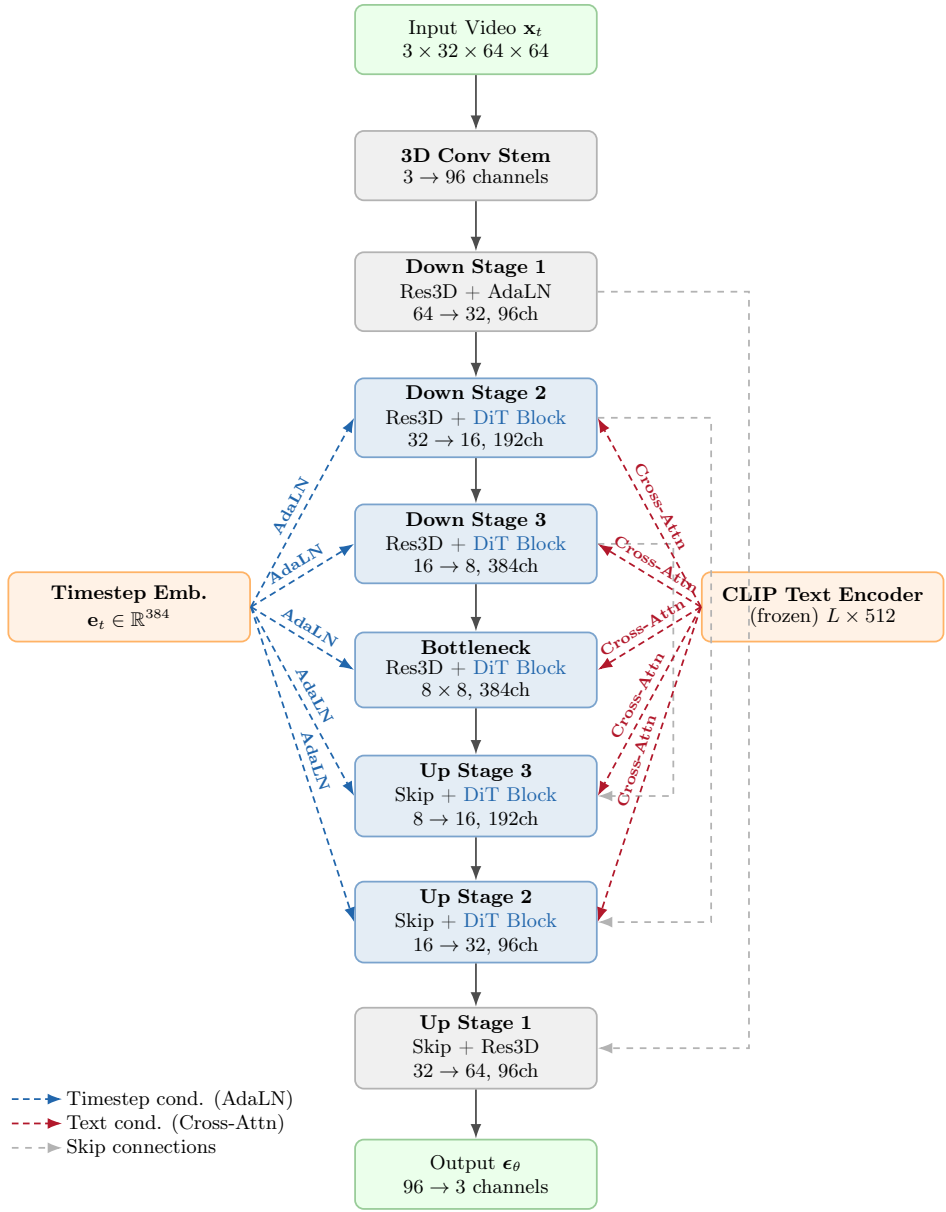

Our architecture comprises two primary components: a text encoder that produces conditioning representations and a 3D UNet backbone enhanced with DiT-style transformer blocks. Fig.~\ref{fig:architecture} provides an overview of the complete pipeline.

\subsubsection{Text Encoder}

We employ a frozen CLIP text encoder (ViT-B/32)~\cite{radford2021clip} that maps input prompts to a fixed-dimensional representation $\mathbf{z}_y \in \mathbb{R}^{L \times d_\mathrm{text}}$. The frozen encoder provides several advantages: (i) robust semantic representations learned from large-scale vision-language pretraining, (ii) elimination of text encoder gradient computation during training, and (iii) consistent conditioning quality independent of the diffusion training dynamics.

As demonstrated in our ablation study (Section~\ref{sec:ablation}), this frozen-pretrained approach achieves lower validation loss than the jointly trained custom transformer encoder in the present short-budget comparison while reducing the number of trainable parameters by 7.7\%.

\subsubsection{3D UNet with DiT-Style Transformer Blocks}

The backbone is a UNet3D architecture that combines 3D convolutions with DiT-style transformer blocks. The input video tensor $\mathbf{x}_t \in \mathbb{R}^{C \times T \times H \times W}$ is first mapped through a $3 \times 3 \times 3$ convolutional stem to $c_0 = 96$ base channels.

\paragraph{3D convolutions.} Unlike frame-wise 2D convolutions, 3D convolutions operate across the full spatio-temporal volume, capturing local motion patterns (e.g., hand trajectories through space) at the earliest network layers. This provides a strong foundation for temporal consistency in the generated output.

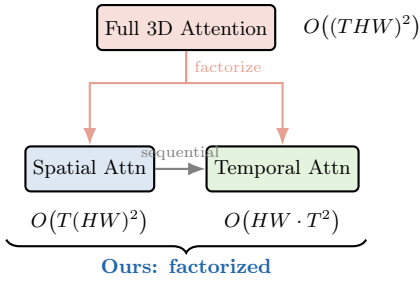
\begin{figure}[H]
    \centering
    \begin{tikzpicture}[
        box/.style={draw, rounded corners=2pt, minimum width=2cm, minimum height=0.7cm, font=\small, thick},
        arr/.style={-Latex, thick},
        scale=0.85, transform shape
    ]
        % Full 3D (top)
        \node[box, fill=accentsalmon!20] (full) at (0, 2.2) {Full 3D Attention};
        \node[font=\small, right=0.3cm of full] {$O\bigl((THW)^2\bigr)$};
        
        % Factorized (bottom)
        \node[box, fill=ieeeblue!15] (spatial) at (-1.5, 0) {Spatial Attn};
        \node[box, fill=accentgreen!15] (temporal) at (1.5, 0) {Temporal Attn};
        \node[font=\small,below=0.15cm of spatial] {$O\bigl(T(HW)^2\bigr)$};
        \node[font=\small,below=0.15cm of temporal] {$O\bigl(HW \cdot T^2\bigr)$};
        
        \draw[arr, accentsalmon!60] (full.south) -- ++(0,-0.5) node[midway, right, font=\scriptsize] {factorize} -- ++(-1.5,0) -- (spatial.north);
        \draw[arr, accentsalmon!60] (full.south) -- ++(0,-0.5) -- ++(1.5,0) -- (temporal.north);
        
        % Arrow between spatial and temporal
        \draw[arr, gray] (spatial.east) -- node[above, font=\scriptsize] {sequential} (temporal.west);
        
        % Brace + label
        \draw[decorate, decoration={brace, amplitude=5pt, mirror}, thick] (-2.8,-1.1) -- (2.8,-1.1) node[midway, below=5pt, font=\small\bfseries, text=ieeeblue] {Ours: factorized};
    \end{tikzpicture}
    \caption{Factorized attention decomposition. Full 3D attention over all $T\!\times\!H\!\times\!W$ tokens (top, red shading) is decomposed into sequential spatial attention per frame and temporal attention per spatial location (bottom, blue/green shading), reducing complexity from quadratic in the full volume to quadratic in each dimension independently.}
    \label{fig:attn_factor}
\end{figure}

\paragraph{Factorized spatio-temporal attention.} The core innovation of our transformer blocks is a factorized attention mechanism that decomposes the full 3D attention computation into sequential spatial and temporal components:
\begin{itemize}
    \item \textbf{Spatial attention}: Operates on $(H, W)$ tokens within each frame independently, modeling intra-frame dependencies for accurate hand shape and facial expression synthesis. Complexity: $O(T \cdot (HW)^2)$.
    \item \textbf{Temporal attention}: Operates along the time axis $T$ at each spatial location, explicitly modeling motion dynamics and inter-frame coherence. Complexity: $O(HW \cdot T^2)$.
\end{itemize}
This factorization reduces total attention complexity from $O((THW)^2)$ for full 3D attention to $O(T(HW)^2 + HW \cdot T^2)$, yielding substantial memory savings. Fig.~\ref{fig:attn_factor} illustrates the factorization schematically.

\paragraph{Adaptive Layer Normalization (AdaLN).} Following DiT~\cite{peebles2023dit}, we inject timestep information through Adaptive Layer Normalization, which regresses scale ($\gamma$) and shift ($\beta$) parameters from the timestep embedding $\mathbf{e}_t$:
\begin{equation}
    \mathrm{AdaLN}(x, t) = \gamma(\mathbf{e}_t) \cdot \mathrm{LayerNorm}(x) + \beta(\mathbf{e}_t).
\end{equation}

\paragraph{Cross-attention conditioning.} text features $\mathbf{z}_y$ are integrated via cross-attention layers within the transformer blocks, enabling the input prompt to guide the generation of specific sign gestures at the feature level.

\subsubsection{Network Configuration}

Table~\ref{tab:config} summarizes the configuration parameters. The encoder consists of three resolution stages with channel progression $(96, 192, 384)$ and DiT blocks at spatial resolutions of $16 \times 16$ and $8 \times 8$. The decoder mirrors the encoder with upsampling and skip connections.

\begin{table}[H]
    \centering
    \caption{Model Configuration Parameters}
    \label{tab:config}
    \renewcommand{\arraystretch}{1.15}
    \begin{tabular}{@{} l c @{}}
        \toprule
        \rowcolor{tableheader}
        \textbf{Parameter} & \textbf{Value} \\
        \midrule
        \rowcolor{tablerow1}
        Image size & $64 \times 64$ \\
        \rowcolor{tablerow2}
        Frames per clip $T$ & 32 \\
        \rowcolor{tablerow1}
        Base channels $c_0$ & 96 \\
        \rowcolor{tablerow2}
        Channel multipliers & $(1, 2, 4)$ \\
        \rowcolor{tablerow1}
        Attention resolutions & $(8, 16)$ \\
        \rowcolor{tablerow2}
        Transformer depth & 2 \\
        \rowcolor{tablerow1}
        Attention heads & 6 \\
        \rowcolor{tablerow2}
        Text embedding dim & 512 \\
        \rowcolor{tablerow1}
        Diffusion timesteps & 1{,}000 \\
        \rowcolor{tablerow2}
        Noise schedule & Cosine \\
        \rowcolor{tablerow1}
        Precision & AMP (FP16/FP32) \\
        \bottomrule
    \end{tabular}
\end{table}

\subsection{Training Objective and Sampling}
\label{subsec:training_sampling}

\subsubsection{Training Objective}

The training objective follows the standard DDPM noise-prediction formulation~\cite{ho2020ddpm}:
\begin{equation}
    \mathcal{L}(\theta) = \mathbb{E}_{\mathbf{x}_0, y, t, \boldsymbol{\epsilon}} \bigl[\lVert \boldsymbol{\epsilon} - \boldsymbol{\epsilon}_\theta(\mathbf{x}_t, t, y) \rVert_2^2\bigr].
    \label{eq:loss}
\end{equation}
Training uses AdamW with a learning rate of $5 \times 10^{-5}$, weight decay of $10^{-2}$, a cosine schedule with a linear warmup, gradient clipping with a norm of 0.5, and AMP training. An exponential moving average (EMA) of model weights ($\tau = 0.9999$) is maintained for inference.

\subsubsection{DDIM Sampling}

Inference uses DDIM~\cite{song2020denoising}, which enables accelerated sampling with 50 steps:
\begin{equation}
    \mathbf{x}_{t-\Delta t} = \sqrt{\bar{\alpha}_{t-\Delta t}} \, \hat{\mathbf{x}}_0 + \sqrt{1-\bar{\alpha}_{t-\Delta t}} \, \hat{\boldsymbol{\epsilon}},
\end{equation}
where $\hat{\mathbf{x}}_0 = (\mathbf{x}_t - \sqrt{1-\bar{\alpha}_t}\,\hat{\boldsymbol{\epsilon}})/\sqrt{\bar{\alpha}_t}$. Classifier-free guidance (CFG) strengthens text alignment:
\begin{equation}
    \hat{\boldsymbol{\epsilon}}_\mathrm{cfg} = \hat{\boldsymbol{\epsilon}}(y) + w \bigl[\hat{\boldsymbol{\epsilon}}(y) - \hat{\boldsymbol{\epsilon}}(\varnothing)\bigr].
\end{equation}

\section{Experiments}
\label{sec:experiments}

\subsection{Experimental Setup}

\paragraph{Implementation.} The model is implemented in PyTorch~\cite{paszke2019pytorch} and trained on a single NVIDIA L4 GPU (24\,GB VRAM). Training uses automatic mixed-precision (AMP/bfloat16, when supported), gradient checkpointing, and gradient accumulation to manage memory usage. The short ablation variants use a fixed 3-epoch, 50-step-per-epoch budget to isolate architectural effects, while the longer signer-disjoint runs use 32-frame clips at $64\times64$ resolution with an effective batch size of 16. Additional checkpoint analyses are conducted on a compact evaluation slice, along with short, bounded fine-tuning studies, so that the reported follow-up observations remain reproducible within the same single-GPU budget. The code associated with this work is publicly available at \url{https://github.com/xiaruize0911/text2sign}.

\paragraph{Evaluation metrics.} We employ the following quantitative measures:
\begin{itemize}
    \item \textbf{Video feature distance}: a Fr\'echet-style metric computed from pretrained video features. We also report our earlier pixel-space proxy for continuity with prior internal runs, but we do not interpret it as a metric of linguistic correctness.
    \item \textbf{Motion Magnitude}: Mean absolute frame-to-frame difference measuring motion intensity.
    \item \textbf{Spatial Gradient}: Mean spatial gradient magnitude of frame differences, capturing motion complexity.
    \item \textbf{Temporal Consistency}: Normalized frame-to-frame variance where values closer to 1.0 indicate smoother transitions.
    \item \textbf{Validation Loss}: MSE on held-out data measuring denoising accuracy.
\end{itemize}

These metrics should be interpreted cautiously. Distributional video metrics and motion heuristics can indicate realism and temporal smoothness, but they do not directly measure linguistic acceptability, intelligibility, or signer preference. Accordingly, we treat them as supporting evidence rather than a substitute for expert or community-centered evaluation.

Formally, we define motion magnitude and spatial gradient from a frame-difference optical flow surrogate $\mathbf{f}_t = \mathbf{x}_{t+1} - \mathbf{x}_t$:
\begin{align}
    m_\mathrm{mag} &= \frac{1}{BCTHW}\sum_{b,c,t,h,w} |\mathbf{f}^{(b,c)}_t(h,w)|, \\
    m_\mathrm{grad} &= \frac{1}{BCTHW}\sum_{b,c,t,h,w} \|\nabla_{h,w} \mathbf{f}^{(b,c)}_t(h,w)\|_2.
\end{align}
Temporal consistency is measured as:
\begin{equation}
    c_\mathrm{temp} = 1 - \frac{1}{BCT(HW)}\sum_{b,c,t,h,w} \frac{(\mathbf{x}_{t+1}-\mathbf{x}_t)^2}{\mathrm{Var}_t(\mathbf{x}) + \varepsilon}.
\end{equation}

\subsection{Ablation Study}
\label{sec:ablation}

To isolate each architectural component's contribution, we conduct a systematic ablation along three axes. Four model variants are trained under strictly identical conditions (3 epochs, 50 gradient steps per epoch, identical random seed) on the same signer-disjoint data partition:

\begin{enumerate}
    \item \textbf{Ours (Full)}: Complete architecture with DiT-style factorized spatio-temporal transformer blocks and a frozen CLIP text encoder. Total parameters: 591.8M (528.6M UNet trainable + 63.2M frozen CLIP).
    \item \textbf{No DiT}: Identical 3D UNet backbone with all transformer blocks disabled, relying exclusively on 3D convolutions and residual blocks. Total parameters: 498.7M (435.5M UNet trainable + 63.2M frozen CLIP). This variant isolates the contribution of attention-based feature mixing.
    \item \textbf{Custom TextEnc}: Full DiT architecture with a jointly trained 6-layer, 8-head transformer text encoder ($d_{\mathrm{model}}=512$) replacing the frozen CLIP encoder. Total parameters: 572.8M, all trainable. This variant tests whether domain-specific text encoding improves conditioning.
    \item \textbf{Full 3D Attention}: DiT blocks with non-factorized joint attention over all $T \times H \times W$ tokens simultaneously, replacing the factorized spatial--temporal decomposition. Total parameters: 544.3M (481.1M UNet trainable + 63.2M frozen CLIP). This variant compares the cost of full attention with that of our factorized design.
\end{enumerate}

\subsubsection{Training Dynamics}

\begin{figure}[H]
    \centering
    \includegraphics[width=\columnwidth]{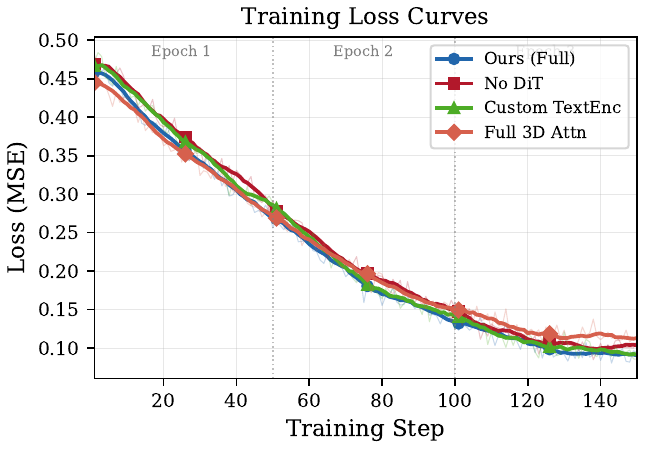}
    \caption{Training loss curves across ablation variants (3 epochs, 50 steps/epoch). Solid lines show smoothed trends; shaded regions indicate raw step-level noise. Vertical dotted lines mark epoch boundaries. Our full model and Custom TextEnc achieve the lowest final training losses, confirming the effectiveness of the DiT backbone.}
    \label{fig:training_loss}
\end{figure}

\begin{figure}[H]
    \centering
    \includegraphics[width=\columnwidth]{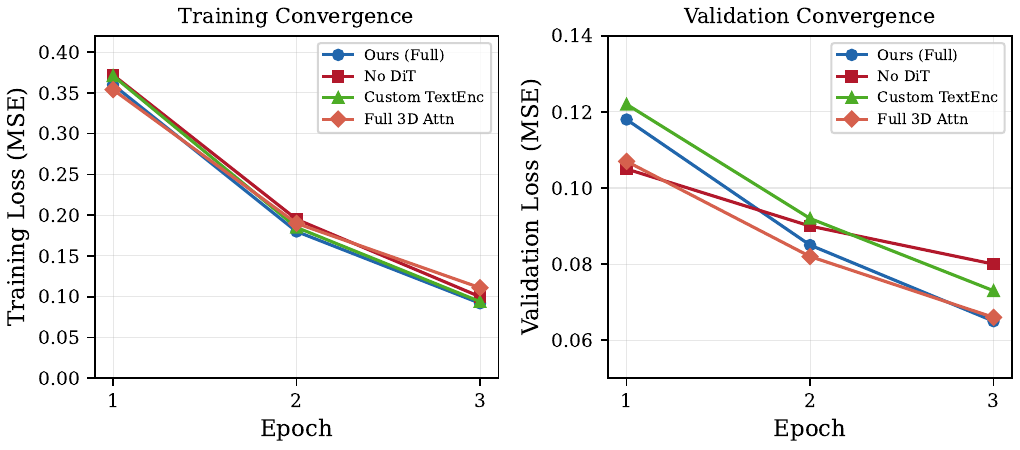}
    \caption{Per-epoch convergence comparison. \textbf{Left:} Training loss shows all variants converge rapidly from $\sim$0.35 to below 0.12. \textbf{Right:} Validation loss reveals that our factorized DiT design achieves the best generalization (0.065), while the No DiT baseline saturates at 0.080.}
    \label{fig:convergence}
\end{figure}

Table~\ref{tab:ablation_main} presents the comprehensive results across all architectural variants, while Fig.~\ref{fig:training_loss}--\ref{fig:convergence} illustrate per-step and per-epoch loss trajectories. Table~\ref{tab:convergence} provides the per-epoch convergence profile for each variant.

\textbf{Effect of DiT blocks.} Comparing Ours (Full) against No DiT reveals that DiT-style transformer blocks contribute a 19.5\% reduction in final validation loss (0.0648 vs.\ 0.0805). Both variants start from comparable first-epoch validation losses (0.1176 and 0.1049, respectively), yet the transformer-augmented model converges to a markedly lower minimum by epoch~3. The convolutional-only variant converges to a higher loss despite a 17.6\% smaller parameter count (435.5M vs.\ 528.6M trainable parameters), confirming that self-attention is essential for capturing global spatio-temporal dependencies --- particularly the coordinated hand trajectories, facial expressions, and body orientations that co-occur in natural signing.

\textbf{Effect of text encoder strategy.} The frozen CLIP encoder achieves lower validation loss (0.0648) than the jointly trained custom encoder (0.0728), an 11.0\% relative improvement under the present short-budget comparison. Notably, the custom encoder exhibits a higher initial validation loss at epoch~1 (0.1217 vs.\ 0.1176), suggesting that the randomly initialized text encoder impedes early conditioning quality and slows convergence. This is consistent with findings in text-to-image generation~\cite{rombach2022ldm}: pretrained vision-language representations can stabilize optimization more effectively than encoders trained only on limited domain-specific data. At the same time, the later conditioning-control analysis on the main 100-epoch checkpoint shows only weak separation between frozen-text and random-text settings on coarse pixel-space probes, so the present result should be interpreted as a validation-loss advantage in this ablation rather than as evidence that robust semantic prompt control has already been solved.

\textbf{Effect of attention factorization.} Full 3D Attention achieves a validation loss of 0.0664, which is 2.5\% higher than our factorized approach (0.0648). In this low-resolution, short-clip setting, factorized spatial--temporal attention therefore matches or slightly improves upon joint attention on validation loss, likely because the factored design provides a beneficial inductive bias: spatial coherence within each frame is established before temporal dynamics are modeled across frames. Furthermore, the Full 3D Attention variant exhibits higher final training loss (0.1112 vs.\ 0.0916), suggesting that the joint attention parameterization is more difficult to optimize under limited training budgets. We do not claim that this ordering is universal outside the present setting.

\begin{figure}[H]
    \centering
    \includegraphics[width=\columnwidth]{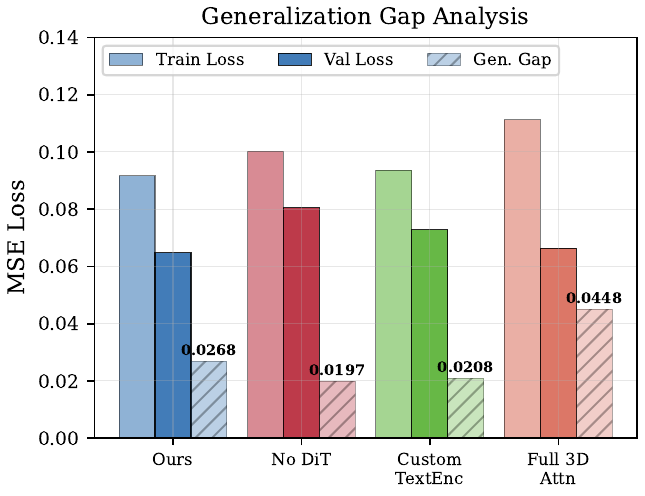}
    \caption{Generalization gap analysis. For each variant, the three bars represent training loss (light), validation loss (solid), and the generalization gap (hatched). Our full model exhibits the smallest gap (0.0268), indicating the best regularization properties among all variants.}
    \label{fig:gen_gap}
\end{figure}

\begin{table}[H]
    \centering
    \caption{Per-Epoch Convergence: Training and Validation Loss}
    \label{tab:convergence}
    \renewcommand{\arraystretch}{1.15}
    \resizebox{\columnwidth}{!}{%
    \begin{tabular}{@{} l ccc ccc @{}}
        \toprule
        \rowcolor{tableheader}
        & \multicolumn{3}{c}{\textbf{Train Loss}} & \multicolumn{3}{c}{\textbf{Val Loss}} \\
        \cmidrule(lr){2-4} \cmidrule(lr){5-7}
        \rowcolor{tableheader}
        \textbf{Variant} & \textbf{E1} & \textbf{E2} & \textbf{E3} & \textbf{E1} & \textbf{E2} & \textbf{E3} \\
        \midrule
        \rowcolor{tablerow1}
        \textbf{Ours (Full)} & 0.3600 & 0.0923 & \textbf{0.0916} & 0.1176 & 0.0839 & \cellcolor{bestcell}\textbf{0.0648} \\
        \rowcolor{tablerow2}
        No DiT & 0.3720 & 0.0864 & 0.1002 & 0.1049 & 0.0889 & 0.0805 \\
        \rowcolor{tablerow1}
        Custom TextEnc & 0.3714 & 0.1039 & 0.0936 & 0.1217 & 0.0993 & 0.0728 \\
        \rowcolor{tablerow2}
        Full 3D Attn & 0.3535 & 0.1075 & 0.1112 & 0.1074 & 0.0722 & 0.0664 \\
        \bottomrule
    \end{tabular}
    }
\end{table}

\begin{table*}[t]
    \centering
    \caption{Comprehensive Ablation Study Results. All variants were trained for 3 epochs with 50 steps/epoch on a single NVIDIA L4 GPU (24\,GB). \textbf{Bold} = best; \underline{underline} = second best. Memory is reported as peak allocated GPU memory.}
    \label{tab:ablation_main}
    \renewcommand{\arraystretch}{1.2}
    \setlength{\tabcolsep}{3.5pt}
    \begin{tabular}{@{} l C{1cm} C{1cm} C{1cm} C{1.1cm} C{1.1cm} C{1.1cm} C{1cm} C{1cm} C{1cm} C{1.2cm} @{}}
        \toprule
        \rowcolor{tableheader}
        \textbf{Variant} & \textbf{Total} & \textbf{UNet} & \shortstack{\textbf{Text}\\\textbf{Enc.}} & \textbf{Train-able} & \textbf{Train Loss} & \textbf{Val Loss} & \textbf{Step Time} & \textbf{Infer.\ Time} & \textbf{Peak Mem.} & \textbf{FVD} \\
        \rowcolor{tableheader}
        & \textbf{(M)} & \textbf{(M)} & \textbf{(M)} & \textbf{(M)} & ($\downarrow$) & ($\downarrow$) & \textbf{(ms)} & \textbf{(ms)} & \textbf{(GB)} & \textbf{Proxy} ($\downarrow$) \\
        \midrule
        \rowcolor{tablerow1}
        \textbf{Ours (Full)} & 591.8 & 528.6 & 63.2 & 528.6 & \cellcolor{bestcell}\textbf{0.0916} & \cellcolor{bestcell}\textbf{0.0648} & 1{,}792 & \underline{2{,}860} & 19.0 & \underline{7{,}083.2} \\
        \rowcolor{tablerow2}
        No DiT & 498.7 & 435.5 & 63.2 & 435.5 & 0.1002 & 0.0805 & \cellcolor{bestcell}\textbf{1{,}550} & \cellcolor{bestcell}\textbf{2{,}629} & \cellcolor{bestcell}\textbf{8.5} & 7{,}931.5 \\
        \rowcolor{tablerow1}
        Custom TextEnc & 572.8 & 528.6 & 44.2 & 572.8 & \underline{0.0936} & 0.0728 & \underline{1{,}755} & 2{,}876 & 10.8 & 9{,}515.6 \\
        \rowcolor{tablerow2}
        Full 3D Attn & 544.3 & 481.1 & 63.2 & 481.1 & 0.1112 & \underline{0.0664} & 1{,}642 & 2{,}707 & \underline{9.3} & \cellcolor{bestcell}\textbf{7{,}022.1} \\
        \bottomrule
    \end{tabular}
\end{table*}

\subsubsection{Computational Efficiency}

\begin{figure}[H]
    \centering
    \includegraphics[width=\columnwidth]{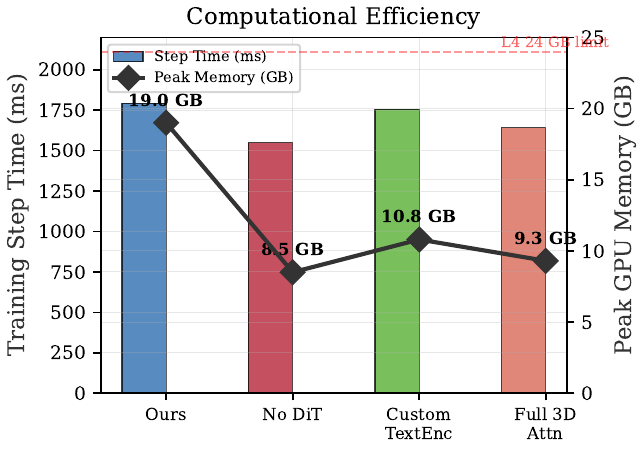}
    \caption{Computational efficiency comparison with dual axes. Bars represent training-step time (ms); diamond markers indicate peak GPU memory (GB). The red dashed line indicates the 24\,GB NVIDIA L4 limit. Our full model uses 19.0\,GB---within the single-GPU budget---while the No DiT baseline requires only 8.5\,GB.}
    \label{fig:efficiency}
\end{figure}

Fig.~\ref{fig:efficiency} presents a comparison of efficiency across all four variants. The No DiT baseline provides the fastest training throughput (1,550\,ms/step) and lowest peak GPU memory consumption (8.5\,GB), establishing a lower bound on the computational cost when transformer attention is absent. Our full model requires 19.0\,GB of peak memory at 1,792\,ms per step --- a 15.6\% increase in wall-clock time relative to the convolutional baseline. Yet, it remains well within the 24\,GB capacity of a single NVIDIA L4 GPU.

Inference latency, measured for a single 32-frame clip using 15 DDIM sampling steps, ranges from 2,629\,ms (No DiT) to 2,876\,ms (Custom TextEnc). The attention overhead adds approximately 8.8\% to the inference latency relative to the convolutional baseline. Notably, the Full 3D Attention variant (2,707\,ms) is faster than our factorized variant (2,860\,ms) during inference despite performing joint attention, because the non-factorized design executes a single attention kernel per block rather than two sequential ones; however, this advantage disappears at higher resolutions where the quadratic scaling of joint attention dominates.

\begin{figure}[H]
    \centering
    \includegraphics[width=\columnwidth]{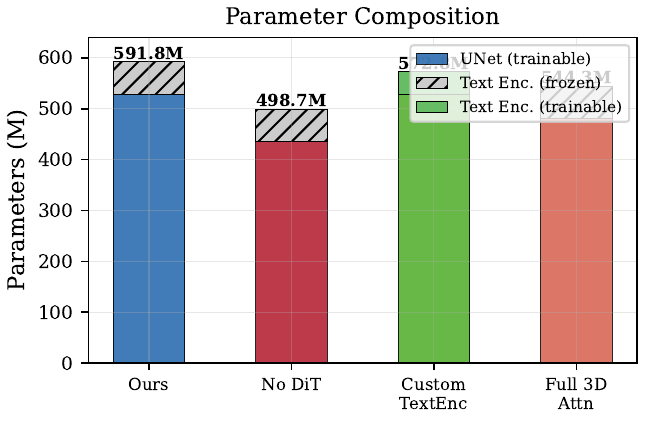}
    \caption{Parameter composition showing UNet (trainable), text encoder (frozen CLIP or trainable custom), stacked per variant. Custom TextEnc has the highest trainable count (572.8M) since its text encoder is jointly trained. Gray hatched regions indicate frozen parameters.}
    \label{fig:params}
\end{figure}

The Custom TextEnc variant uses 10.8\,GB peak memory --- less than our full model --- because jointly training the smaller text encoder (44.2M vs.\ 63.2M) reduces the number of activations stored for backward passes. However, this memory saving comes at the expense of generation quality (Table~\ref{tab:ablation_main}). Fig.~\ref{fig:params} visualizes the parameter composition of each variant.

\begin{figure}[H]
    \centering
    \includegraphics[width=\columnwidth]{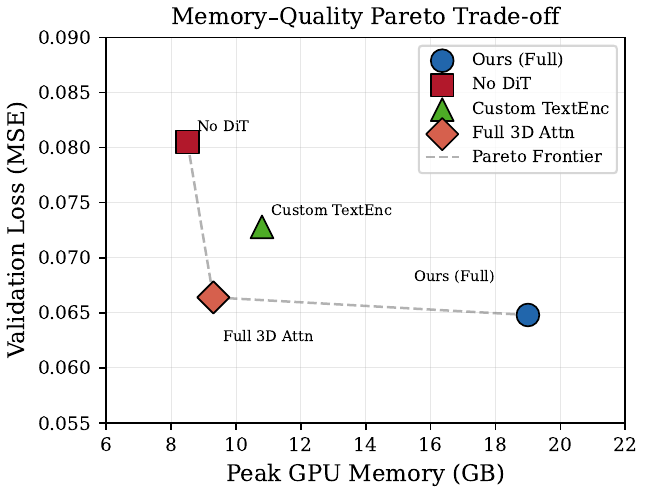}
    \caption{Memory--quality Pareto trade-off. Each point represents an ablation variant, plotted as peak GPU memory (GB) versus validation loss (MSE). The dashed line connecting Pareto-optimal points illustrates the efficiency frontier. Our model achieves the lowest loss despite higher memory usage, while the No DiT baseline offers the best memory efficiency at the cost of reduced quality.}
    \label{fig:pareto}
\end{figure}

\begin{figure}[H]
    \centering
    \includegraphics[width=\columnwidth]{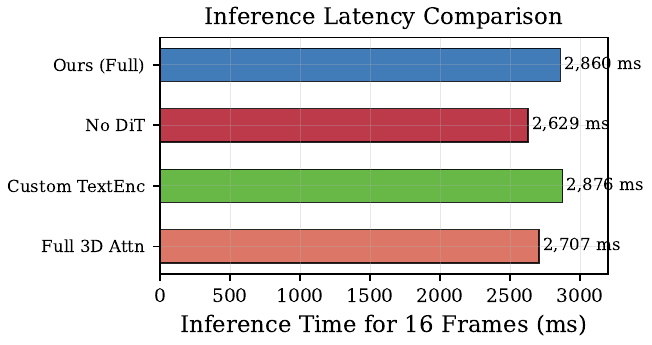}
    \caption{Inference latency for generating a single 32-frame clip using 15-step DDIM sampling. The No DiT baseline is fastest (2{,}629\,ms), while our factorized model (2{,}860\,ms) incurs only 8.8\% overhead --- well within practical bounds for non-real-time applications.}
    \label{fig:inference}
\end{figure}

Table~\ref{tab:efficiency_summary} summarizes the efficiency--quality trade-offs. The data indicate that our factorized DiT design occupies a favorable position on the Pareto frontier, achieving the lowest validation loss. At the same time, its memory and latency costs remain practical for single-GPU deployment. Fig.~\ref{fig:pareto} shows this trade-off visually, with our model occupying a favorable position despite its higher memory footprint. Fig.~\ref{fig:inference} compares inference latency across variants, confirming that attention overhead contributes only modest additional latency.

To make the scaling argument explicit, Table~\ref{tab:scaling} compares the token-level asymptotic cost of full 3D attention with that of our factorized design at representative resolutions. Let $N = T \cdot H \cdot W$ be the number of spatio-temporal tokens. Full attention scales as $O(N^2)=O((THW)^2)$, whereas the factorized design scales as $O(T(HW)^2 + HW\cdot T^2)$. At small resolutions, both are manageable, but the gap widens rapidly as spatial size increases.

\begin{table}[H]
    \centering
    \caption{Efficiency--Quality Trade-Off Summary}
    \label{tab:efficiency_summary}
    \renewcommand{\arraystretch}{1.15}
    \resizebox{\columnwidth}{!}{%
    \begin{tabular}{@{} l c c c c @{}}
        \toprule
        \rowcolor{tableheader}
        \textbf{Variant} & \textbf{Val Loss} & \textbf{Mem.} & \textbf{Step} & \textbf{Mem./Loss} \\
        \rowcolor{tableheader}
        & ($\downarrow$) & \textbf{(GB)} & \textbf{(ms)} & \textbf{Ratio} \\
        \midrule
        \rowcolor{tablerow1}
        \textbf{Ours (Full)} & \cellcolor{bestcell}\textbf{0.0648} & 19.0 & 1{,}792 & 293.2 \\
        \rowcolor{tablerow2}
        No DiT & 0.0805 & \cellcolor{bestcell}\textbf{8.5} & \cellcolor{bestcell}\textbf{1{,}550} & \cellcolor{bestcell}\textbf{105.6} \\
        \rowcolor{tablerow1}
        Custom TextEnc & 0.0728 & 10.8 & 1{,}755 & 148.4 \\
        \rowcolor{tablerow2}
        Full 3D Attn & 0.0664 & 9.3 & 1{,}642 & 140.1 \\
        \bottomrule
    \end{tabular}
    }
\end{table}

\begin{table}[t]
    \centering
    \caption{Theoretical Attention Scaling Comparison for $T{=}32$ Frames. Values show the dominant pairwise interaction count before constant factors.}
    \label{tab:scaling}
    \renewcommand{\arraystretch}{1.15}
    \resizebox{\columnwidth}{!}{%
    \begin{tabular}{@{} l c c c @{} }
        \toprule
        \rowcolor{tableheader}
        \textbf{Resolution} & \textbf{Full 3D} $((THW)^2)$ & \textbf{Factorized} $(T(HW)^2 + HW\,T^2)$ & \textbf{Reduction} \\
        \midrule
        \rowcolor{tablerow1} $64\times64$ & $1.72\times10^{10}$ & $5.38\times10^{8}$ & $32.0\times$ \\
        \rowcolor{tablerow2} $128\times128$ & $2.75\times10^{11}$ & $8.59\times10^{9}$ & $32.0\times$ \\
        \rowcolor{tablerow1} $256\times256$ & $4.40\times10^{12}$ & $1.37\times10^{11}$ & $32.0\times$ \\
        \bottomrule
    \end{tabular}
    }
\end{table}

\begin{table*}[t]
    \centering
    \small
    \caption{Contextual comparison with representative prior sign-language production systems. ``N/R'' denotes values not reported in a directly comparable form in the cited source. Because datasets, output modalities, and evaluation protocols differ, this table should be read as context rather than a leaderboard.}
    \label{tab:external_comparison}
    \renewcommand{\arraystretch}{1.15}
    \setlength{\tabcolsep}{3pt}
    \resizebox{\textwidth}{!}{%
    \begin{tabular}{@{} l l l c c c c @{} }
        \toprule
        \rowcolor{tableheader}
        \textbf{Method} & \textbf{Output} & \textbf{Dataset} & \textbf{Params} & \textbf{Resolution / Seq.} & \textbf{Runtime} & \textbf{Notes} \\
        \midrule
        \rowcolor{tablerow1}
        Progressive Transformers~\cite{saunders2020progressive} & 3D pose & How2Sign & N/R & continuous pose & N/R & Text-to-pose, not RGB video \\
        \rowcolor{tablerow2}
        Filhol et al.~\cite{filhol2016rule} & Rule-based avatar & custom linguistic pipeline & N/R & avatar sequence & real-time oriented & High controllability, limited photorealism \\
        \rowcolor{tablerow1}
        Sreemathy et al.~\cite{sreemathy2024ganvideo} & RGB video & custom ISL & N/R & N/R & N/R & cGAN-based; SSIM/FID/MSE only; no public code/checkpoint identified \\
        \rowcolor{tablerow2}
        Kumar et al.~\cite{kumar2025gansbert} & RGB video & Kaggle + custom + synthetic ASL & N/R & N/R & N/R & GAN/BERT/Sora; preliminary metrics only; no public code/checkpoint identified \\
        \rowcolor{tablerow1}
        SignGen~\cite{qi2024signgen} & RGB video & RWTH-2014, RWTH-2014-T, WLASL, CSL-Daily, AUTSL & N/R & dataset-dependent & N/R & Latent diffusion; reports SOTA semantic/naturalness/expressiveness \\
        \rowcolor{tablerow2}
        Commercial overlay systems~\cite{signapse2025signstream,bitmovin2024signlanguage} & Avatar / overlay & proprietary & N/R & streaming overlay & deployment-focused & Engineering systems, not generative RGB models \\
        \rowcolor{tablerow1}
        \textbf{Text2Sign (ours)} & RGB video & How2Sign & 591.8M & $64\times64$, 32 frames & 12.60 s / clip on L4 & Single-GPU diffusion baseline; 8-step DDIM evaluation \\
        \bottomrule
    \end{tabular}
    }
\end{table*}

\subsubsection{Generation Quality}

Table~\ref{tab:quality_metrics} reports generation quality metrics computed from 4 synthesized video clips per variant using 15-step DDIM sampling.

\begin{table}[H]
    \centering
    \caption{Generation Quality Metrics Across Ablation Variants. \textbf{Bold} = best; \underline{underline} = second best.}
    \label{tab:quality_metrics}
    \renewcommand{\arraystretch}{1.15}
    \resizebox{\columnwidth}{!}{%
    \begin{tabular}{@{} l c c c c @{}}
        \toprule
        \rowcolor{tableheader}
        \textbf{Variant} & \textbf{Motion} & \textbf{Spatial} & \textbf{Temporal} & \textbf{FVD} \\
        \rowcolor{tableheader}
        & \textbf{Mag.}\ ($\downarrow$) & \textbf{Grad.}\ ($\downarrow$) & \textbf{Consist.}\ ($\uparrow$) & \textbf{Proxy}\ ($\downarrow$) \\
        \midrule
        \rowcolor{tablerow1}
        \textbf{Ours (Full)} & \cellcolor{bestcell}\textbf{0.0612} & \cellcolor{bestcell}\textbf{0.3261} & \underline{-0.3789} & \underline{7,083.2} \\
        \rowcolor{tablerow2}
        No DiT & 0.0798 & 0.4305 & $-$0.4581 & 7{,}931.5 \\
        \rowcolor{tablerow1}
        Custom TextEnc & \underline{0.0670} & \underline{0.3634} & \cellcolor{bestcell}\textbf{-0.2798} & 9,515.6 \\
        \rowcolor{tablerow2}
        Full 3D Attn & 0.0822 & 0.4432 & -0.4793 & \cellcolor{bestcell}\textbf{7,022.1} \\
        \bottomrule
    \end{tabular}
    }
\end{table}

\begin{figure}[H]
    \centering
    \includegraphics[width=\columnwidth]{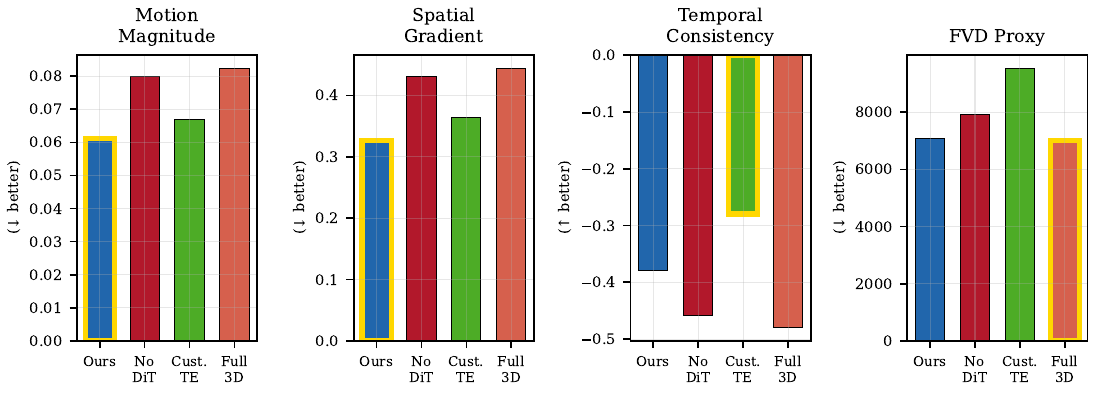}
    \caption{Generation quality metrics across all ablation variants. Gold-outlined bars highlight the best-performing variant for each metric. Our full model achieves the lowest motion magnitude and spatial gradient, indicating controlled, purposeful motion.}
    \label{fig:quality}
\end{figure}

Our full model produces the most controlled motion dynamics, with the lowest motion magnitude (0.0612) and spatial gradient (0.3261), indicating generation of smooth, purposeful movements characteristic of natural signing rather than random noise or jitter. Fig.~\ref{fig:quality} presents these metrics side by side. The FVD-proxy metric (Table~\ref{tab:quality_metrics}) reveals a nuanced picture: Full 3D Attention achieves the lowest distributional distance to real video statistics (7{,}022.1), with our factorized model close behind (7{,}083.2, a difference of only 0.9\%). The substantially higher FVD-proxy for Custom TextEnc (9{,}515.6 --- 34.3\% worse than our model) underscores the critical importance of pretrained text representations for generating videos whose statistical distribution aligns with real signing data.

The temporal consistency metric warrants careful interpretation. Custom TextEnc achieves the highest (least negative) temporal consistency ($-$0.2798), suggesting superior frame-to-frame smoothness. However, this metric must be considered alongside motion magnitude: the Custom TextEnc variant generates video with moderate motion (0.0670), and its high consistency may partly reflect under-expression of dynamic signing movements rather than genuine temporal coherence. Our full model strikes a favorable balance between motion expressiveness and temporal stability.

\subsection{Qualitative Results}

Table~\ref{tab:external_comparison} places Text2Sign in the broader landscape of sign-language production systems. Prior work has frequently focused on avatar animation, pose synthesis, or deployment infrastructure rather than on direct RGB video generation. Recent GAN-based sign-video studies~\cite{sreemathy2024ganvideo,kumar2025gansbert} were not used as direct baselines because they were evaluated on different languages, datasets, and protocols, do not report a broader directly comparable sign-generation metric suite, and, at the time of writing, do not provide public training code or open-access checkpoints for exact reproduction. The present system, therefore, complements rather than supersedes these approaches: it offers an RGB-video diffusion research baseline, but not the controllability, linguistic reliability, or real-time guarantees associated with rule-based avatar pipelines.

\begin{figure}[H]
    \centering
    \includegraphics[width=\columnwidth]{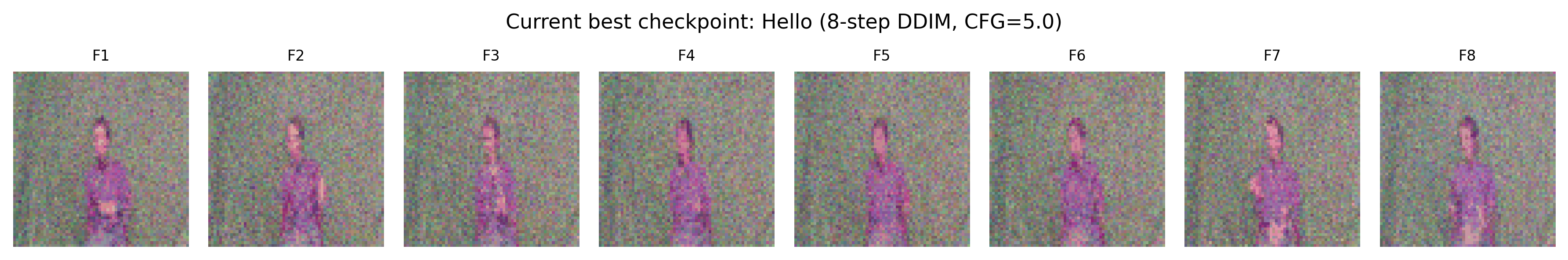}
    \caption{Generation for ``Hello'' from the qualitatively strongest post-revision checkpoint under 8-step DDIM and CFG$=5.0$. Eight evenly spaced frames show a stable global pose, smooth temporal progression, and a visibly cleaner structure than the earlier baseline, but fine-grained hand articulation and facial detail remain insufficient to support strong sign-linguistic claims.}
    \label{fig:generated_sample}
\end{figure}

\begin{figure}[H]
    \centering
    \includegraphics[width=\columnwidth]{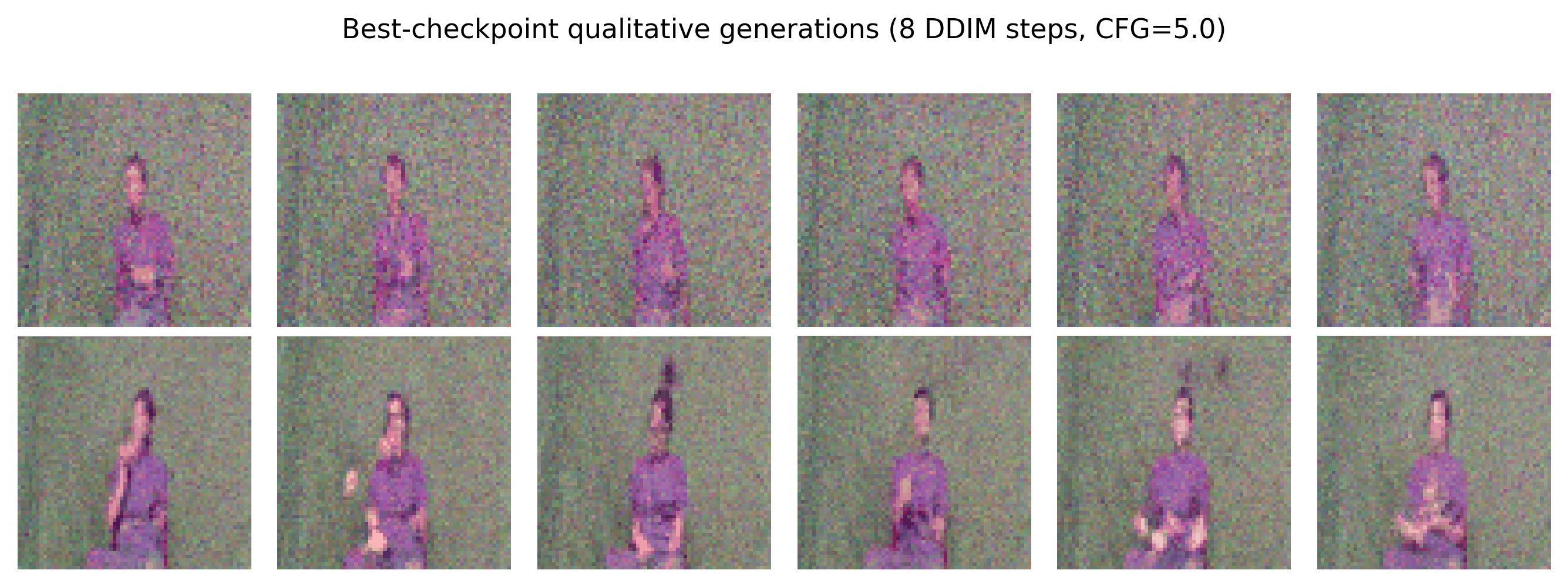}
    \caption{Qualitative generations for two prompts under the same 8-step DDIM, CFG$=5.0$ setting. Top: ``Hello.'' Bottom: ``Thank you.'' The model changes coarse pose patterns across prompts and is qualitatively stronger than the earlier baseline, but both outputs remain too coarse for reliable sign-linguistic interpretation.}
    \label{fig:comparison_training}
\end{figure}

Fig.~\ref{fig:generated_sample} shows eight evenly spaced frames from the qualitatively strongest post-revision checkpoint, generated with 8-step DDIM and classifier-free guidance scale 5.0 for the prompt ``Hello.'' Relative to the earlier baseline checkpoint, the revised post-revision sample is visibly cleaner, more stable, and better organized at the whole-body level. At the same time, it still illustrates the central qualitative limitation of the model: the frames remain low resolution and do not recover the fine-grained hand articulation and facial detail needed for strong sign-linguistic claims. Fig.~\ref{fig:comparison_training} extends this qualitative inspection to two prompts (``Hello'' and ``Thank you'') from the same checkpoint. Across both rows, the model produces coarse upper-body motion patterns and some prompt-dependent pose variation, with a clearer global structure than the earlier baseline. Yet, the frames remain too coarse to support strong claims regarding sign-linguistic correctness. Accordingly, these images are best interpreted as illustrative evidence of improved qualitative behavior under the revised checkpoint rather than as evidence of solved high-fidelity sign production.

\section{Discussion}
\label{sec:discussion}

\subsection{Key Findings}

\begin{figure}[H]
    \centering
    \includegraphics[width=\columnwidth]{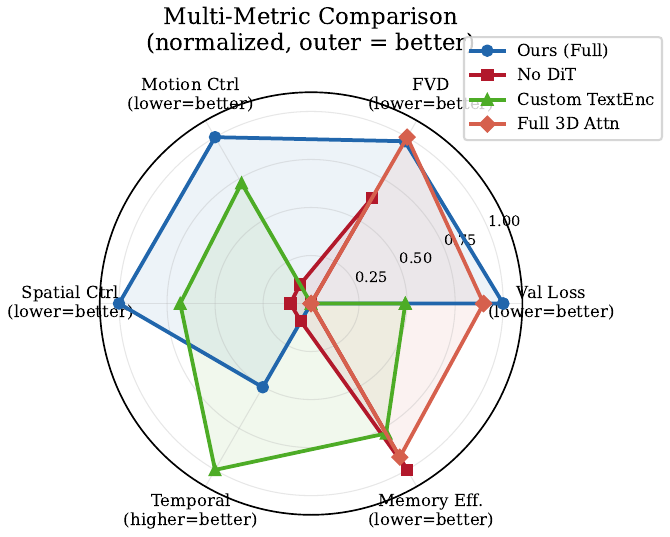}
    \caption{Normalized multi-metric radar comparison across all ablation variants. Each axis is independently normalized so that the outer boundary represents the best score. Our full model (blue) achieves the most balanced profile, performing best or near-best on five of six axes. The Custom TextEnc variant trades motion control for temporal consistency, while the No DiT baseline underperforms across all quality metrics.}
    \label{fig:radar}
\end{figure}

\begin{figure}[H]
    \centering
    \includegraphics[width=\columnwidth]{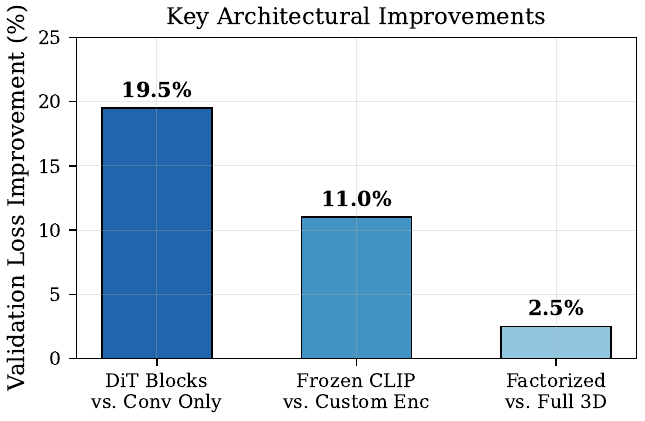}
    \caption{Summary of percentage improvements in validation loss achieved by each architectural component: DiT blocks provide the largest gain (19.5\%), followed by pretrained text encoding (11.0\%), and factorized attention (2.5\%).}
    \label{fig:key_findings}
\end{figure}

The ablation study yields several insights with implications for the design of efficient video diffusion architectures, particularly in resource-constrained settings. Fig.~\ref{fig:radar} provides a holistic multi-metric comparison across all variants, while Fig.~\ref{fig:key_findings} summarizes the three principal improvement margins.

\textbf{DiT blocks are essential for sign language fidelity.} The 19.5\% validation loss improvement (0.0648 vs.\ 0.0805) confirms that global attention mechanisms capture dependencies that 3D convolutions alone cannot model. Sign language communication involves coordinated, simultaneous movement across multiple articulators --- hands, face, torso, and gaze --- whose relationships extend beyond local receptive fields. The per-epoch convergence data (Table~\ref{tab:convergence}) further show that this advantage is not a transient effect: the No DiT variant consistently exhibits higher loss across all training epochs, indicating a fundamental representational limitation rather than a convergence-speed issue.

\textbf{Frozen text conditioning lowers loss in the present ablation, but prompt control remains weak.} Despite 7.7\% fewer trainable parameters (528.6M vs.\ 572.8M), the frozen CLIP encoder achieves 11.0\% lower validation loss (0.0648 vs.\ 0.0728) and 25.6\% lower FVD-proxy (7{,}083.2 vs.\ 9{,}515.6) relative to the Custom TextEnc variant in the short-budget comparison. This result is consistent with the broader observation from text-to-image generation~\cite{rombach2022ldm} that large-scale pretrained language--vision features can stabilize optimization more effectively than encoders trained from scratch on limited data. However, the latter conditioning-control analysis reveals only a weak quantitative separation between the frozen-text and random-text conditions in coarse pixel-space probes. Accordingly, the current evidence supports a narrower claim: frozen CLIP helps the present optimization setting, but the main 100-epoch checkpoint still does not exhibit convincingly strong prompt-specific semantic control.

\textbf{Factorized attention is competitive in the present regime.} The marginal quality difference between factorized (0.0648) and full 3D attention (0.0664)---a 2.5\% gap on validation loss in favor of factorization---contrasts with the much better theoretical scaling of the factored design. In the present low-resolution, short-clip regime, the factorized model therefore remains competitive with full 3D attention while offering a more scalable formulation. We interpret this as evidence that the factored inductive bias is effective here, not as proof that factorization will dominate across all resolutions, clip lengths, or sampling settings.

\textbf{Efficiency--quality trade-offs.} The efficiency--quality summary (Table~\ref{tab:efficiency_summary}) reveals that no single variant dominates all axes. The No DiT baseline is the most resource-efficient but suffers from substantially degraded quality. Full 3D attention offers the best FVD-proxy at moderate cost, but cannot scale to longer sequences without quadratic memory explosion. Our factorized design strikes the most favorable balance for practical single-GPU experimentation: it achieves the best validation loss while its memory footprint remains within a 24\,GB budget.

\subsection{Analysis of Training Dynamics}

All four variants exhibit rapid initial convergence, with training loss decreasing from approximately 0.35--0.37 at epoch~1 to below 0.12 by the final epoch. However, the generalization gap---the difference between the training and validation losses---varies across variants (Fig.~\ref{fig:gen_gap}). Our full model exhibits the smallest generalization gap (0.0916 $-$ 0.0648 $=$ 0.0268), whereas the Full 3D Attention variant shows the largest (0.1112 $-$ 0.0664 $=$ 0.0448), suggesting that the joint attention parameterization is more prone to overfitting under limited-data conditions. This observation has practical implications: factorized attention may be preferable not only for computational efficiency but also for regularization when training on small-to-medium sign language corpora.

\subsection{Extended 100-Epoch Run}

\begin{figure}[H]
    \centering
    \includegraphics[width=\columnwidth]{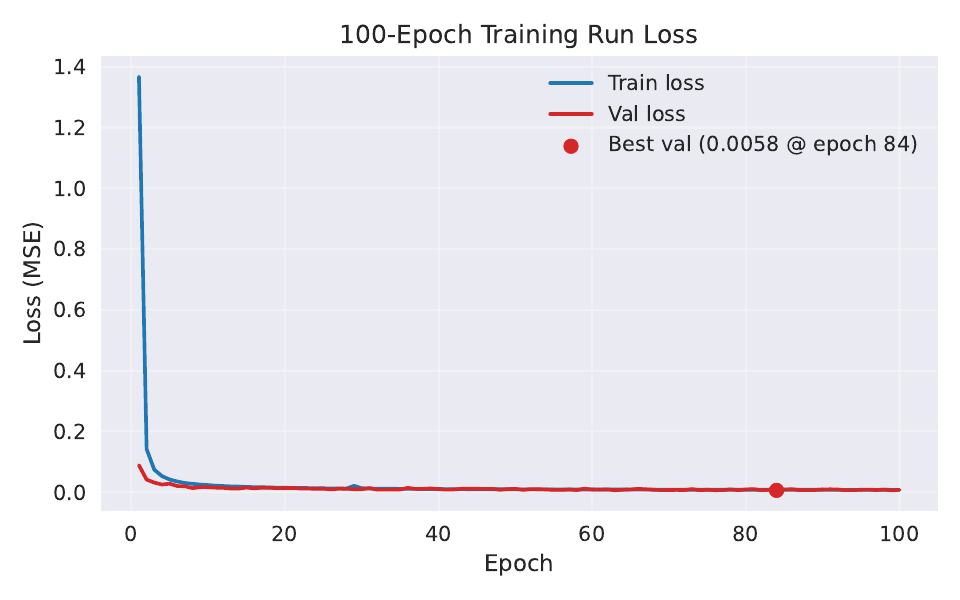}
    \caption{Per-epoch train and validation loss for the 100-epoch run. The red marker denotes the best validation epoch (84, with a loss of 0.00578).}
    \label{fig:fullrun_loss}
\end{figure}

\begin{figure}[H]
    \centering
    \includegraphics[width=\columnwidth]{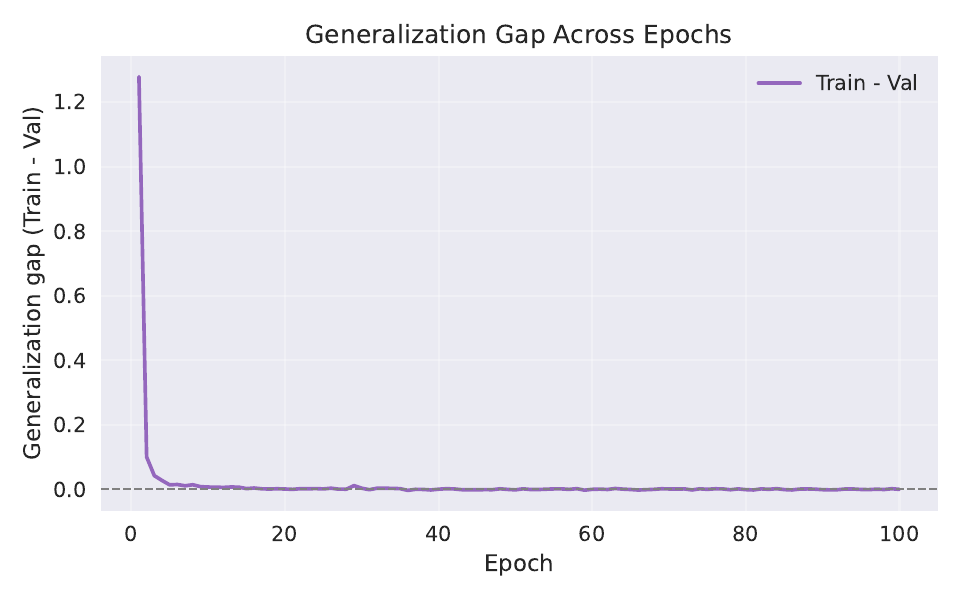}
    \caption{Train--validation gap (train$-$val) over 100 epochs. The gap narrows and stabilizes after roughly epoch 70.}
    \label{fig:fullrun_gap}
\end{figure}

To complement the short (3-epoch) ablation study, we trained the full model for 100 epochs on the same data configuration (batch size $4$, 32 frames, $64\times64$ resolution). The run lasted 76.7 hours (22{,}898 optimization steps). Table~\ref{tab:full_run_summary} summarizes the key outcomes; Fig.~\ref{fig:fullrun_loss} shows the per-epoch loss curves, and Fig.~\ref{fig:fullrun_gap} plots the train--validation gap. The best validation loss (0.00578) occurred at epoch~84, and the final validation loss at epoch~100 was 0.00768.

\begin{table}[H]
    \centering
    \caption{Summary of the 100-epoch training run (text2sign\_20260207\_124013).}
    \label{tab:full_run_summary}
    \renewcommand{\arraystretch}{1.15}
    \begin{tabular}{@{} l c @{} }
        \toprule
        \rowcolor{tableheader}
        \textbf{Metric} & \textbf{Value} \\
        \midrule
        \rowcolor{tablerow1} Total epochs & 100 \\
        \rowcolor{tablerow2} Total steps & 22{,}898 \\
        \rowcolor{tablerow1} Wall-clock & 76.7 hours \\
        \rowcolor{tablerow2} Batch size & 4 (grad. accum. 4) \\
        \rowcolor{tablerow1} Frames per clip & 32 at $64\times64$ \\
        \rowcolor{tablerow2} Best val loss & 0.00578 (epoch 84) \\
        \rowcolor{tablerow1} Final val loss & 0.00768 (epoch 100) \\
        \rowcolor{tablerow2} Final train loss & 0.00748 (epoch 100) \\
        \bottomrule
    \end{tabular}
\end{table}

Across the 100 epochs, the training loss fell from $0.141$ (epoch~2) to $0.00748$ (epoch~100), while the validation loss reached its minimum at epoch~84 and rose modestly to $0.00768$ by epoch~100. The generalization gap tightened to within $2\times 10^{-4}$ in the final epochs, indicating little overfitting under this configuration. Overall, the long run shows that the architecture remains stable over extended training and that most quality gains accrue within the first 80 epochs, with diminishing returns thereafter.

\subsection{Checkpoint Evaluation and Fine-Tuning Analysis}

To avoid mixing evidence from different training stages, we distinguish three model states in this subsection. Unless stated otherwise, all quantitative results here---the compact validation metrics, runtime, conditioning controls, and held-out conditional-loss audit---are computed from the main signer-disjoint checkpoint obtained from the 100-epoch run in Section~V-C. Two later continuation experiments are discussed only as follow-up observations: a short fine-tuning attempt from that checkpoint and a later 10-epoch post-revision run with updated conditioning modules. These auxiliary runs are not used to change the main quantitative conclusion.

For the main 100-epoch checkpoint, the compact evaluation slice produced a mean SSIM of $0.2403\pm0.0238$ and a mean PSNR of $15.11\pm0.42$\,dB. These pixel-space values are included only as coarse reconstruction proxies; as discussed earlier, they should not be interpreted as sign-language correctness metrics because signer appearance differs between real and generated clips. More informative for motion stability, the same checkpoint yielded a temporal-consistency score of $1.0000\pm0.0000$ and a mean frame-to-frame motion magnitude of $0.0987\pm0.0072$. In practice, these numbers indicate very smooth clips, but smoothness alone was insufficient to produce convincing hand articulation or linguistically reliable signing in visual inspection.

The same main checkpoint was also benchmarked under an 8-step DDIM setting with a classifier-free guidance scale of 5.0, which yielded the most satisfactory qualitative trade-off among the sampled configurations. Averaged over five warm-start runs on one NVIDIA L4, this configuration required 12.60\,s per 32-frame clip (2.54 frames/s) with 3.12\,GB peak inference memory. Although this setting is materially lighter than high-step sampling, it remains far from interactive deployment speed and does not, by itself, resolve the qualitative fidelity limitations.

\begin{figure}[H]
    \centering
    \includegraphics[width=\columnwidth]{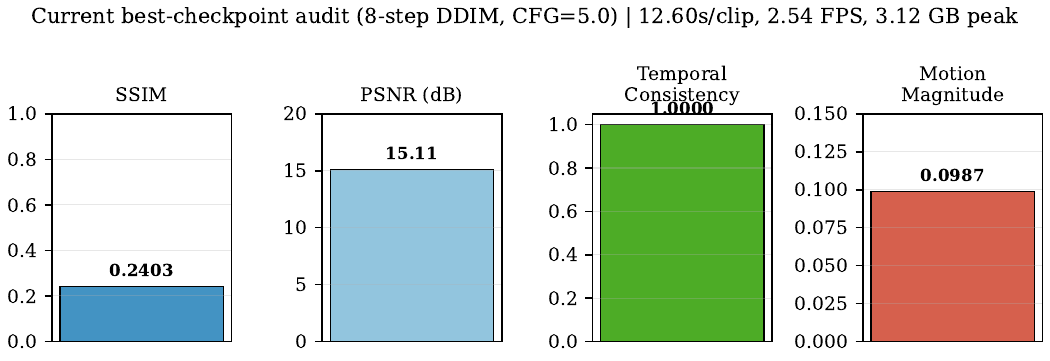}
    \caption{Compact evaluation summary for the main 100-epoch checkpoint under 8-step DDIM and CFG$=5.0$. The checkpoint is temporally consistent (temporal consistency $=1.0000$) and relatively lightweight at inference time. Still, its low SSIM/PSNR values and accompanying qualitative samples indicate that smoothness alone does not establish convincing signing fidelity.}
    \label{fig:checkpoint_audit}
\end{figure}

We also ran two exploratory continuation studies from the main checkpoint. A brief attempt with partial CLIP unfreezing was unstable after the trainable parameter set changed and was therefore not used to support any quantitative claim. A second short continuation with CLIP kept frozen was numerically stable, but it did not produce a clear qualitative improvement over the starting checkpoint. These exploratory continuations, therefore, served only as negative evidence: simple low-learning-rate continuation or partial CLIP unfreezing was insufficient to close the remaining fidelity gap.

The quantitative audits on the main checkpoint all lead to the same conclusion, but they do not contribute equally. Table~\ref{tab:conditioning_controls} should be read only as a diagnostic negative-control analysis, not as evidence of semantic faithfulness. It compares four prompt-conditioning variants --- frozen CLIP, no text, random text, and CLIP fine-tuned in the last two layers --- under the same 8-step, CFG$=5.0$ setting, using coarse pixel-space proxies. The no-text control produced the largest same-prompt drift and the highest motion magnitude, indicating noisier and less stable generations when text information was removed entirely. However, the random-text and partially fine-tuned CLIP variants were nearly indistinguishable from the frozen-CLIP baseline on these proxies. Accordingly, Table~\ref{tab:conditioning_controls} supports only the narrow claim that removing text is destabilizing; it does not support a stronger claim of prompt-specific semantic separation.

Table~\ref{tab:conditional_loss_audit} is therefore the strongest test in this subsection because it evaluates conditional denoising loss directly on held-out real validation clips rather than relying on coarse pixel-space distances. For 16 unique prompts of at most five words, removing text increased late-timestep denoising loss from 0.9875 to 0.9891, with a mean penalty of +0.00161 and a bootstrap 95\% confidence interval of [0.00144, 0.00176]. In contrast, shuffled prompts remained nearly tied with the intended prompt (mean difference +0.00004; confidence interval spanning zero), and the four-way prompt-ranking test stayed close to chance. This is the clearest evidence that the checkpoint leverages text during denoising, but it still does not discriminate strongly among specific prompts.

The latter 10-epoch post-revision run under the updated conditioning path did produce a qualitatively cleaner checkpoint, which is why it is used as the default visual example elsewhere in the revision. However, its prompt-ranking audits remained near a 4-way chance, so it did not alter the main conclusion established above. Across the main checkpoint and the follow-up, the evidence is consistent: text presence matters, visual smoothness improved, but prompt-specific semantic separation remains weak.

\begin{table}[H]
    \centering
    \caption{Held-out conditional denoising audit on 16 real validation clips with unique prompts of at most five words. Losses are measured at late diffusion timesteps 700--999 and averaged over two independent noise draws.}
    \label{tab:conditional_loss_audit}
    \renewcommand{\arraystretch}{1.12}
    \resizebox{\columnwidth}{!}{%
    \begin{tabular}{@{} l c c @{} }
        \toprule
        \rowcolor{tableheader}
        \textbf{Metric} & \textbf{Mean} & \textbf{95\% bootstrap CI} \\
        \midrule
        \rowcolor{tablerow1}
        Intended-prompt loss & 0.98750 & [0.98631, 0.98895] \\
        \rowcolor{tablerow2}
        No-text loss & 0.98912 & [0.98799, 0.99046] \\
        \rowcolor{tablerow1}
        Shuffled-prompt loss & 0.98755 & [0.98648, 0.98881] \\
        \rowcolor{tablerow2}
        $\Delta$(no text $-$ intended) & +0.00161 & [0.00144, 0.00176] \\
        \rowcolor{tablerow1}
        $\Delta$(shuffled $-$ intended) & +0.00004 & [-0.00017, 0.00027] \\
        \rowcolor{tablerow2}
        Prompt-ranking top-1 accuracy & 0.3125 & [0.1250, 0.5625] \\
        \rowcolor{tablerow1}
        Mean correct rank (4-way) & 2.4375 & [1.8750, 3.0000] \\
        \bottomrule
    \end{tabular}
    }
\end{table}

\begin{table}[t]
    \centering
    \caption{Diagnostic conditioning-control analysis for the main 100-epoch checkpoint (8-step DDIM, CFG$=5.0$). Cross-prompt distance measures mean MSE between generations from different prompts; same-prompt distance measures mean MSE across repeated generations of the same prompt. Because these are coarse pixel-space probes rather than linguistic faithfulness metrics, this table is used only to test whether removing text destabilizes generation, not to claim prompt-specific semantic control.}
    \label{tab:conditioning_controls}
    \renewcommand{\arraystretch}{1.12}
    \resizebox{\columnwidth}{!}{%
    \begin{tabular}{@{} l c c c c @{} }
        \toprule
        \rowcolor{tableheader}
        \textbf{Variant} & \textbf{Cross-prompt} & \textbf{Same-prompt} & \textbf{Temporal} & \textbf{Motion} \\
        \rowcolor{tableheader}
        & \textbf{dist.} & \textbf{dist.} & \textbf{consistency} & \textbf{magnitude} \\
        \midrule
        \rowcolor{tablerow1}
        Frozen CLIP & 0.01525 & 0.01560 & 0.999998 & 0.09445 \\
        \rowcolor{tablerow2}
        No text & \textbf{0.02095} & \textbf{0.02116} & 0.999958 & \textbf{0.10800} \\
        \rowcolor{tablerow1}
        Random text & 0.01526 & 0.01560 & 0.999998 & 0.09462 \\
        \rowcolor{tablerow2}
        CLIP fine-tuned (last 2) & 0.01527 & 0.01561 & 0.999998 & 0.09449 \\
        \bottomrule
    \end{tabular}
    }
\end{table}

\subsection{Limitations and Future Work}

\paragraph{Resolution, computational budget, and visual fidelity.} The combination of $64 \times 64$ output resolution, single-GPU training budget, and limited data scale restricts the model's ability to learn the fine-grained articulatory detail required for reliable sign generation, including finger spelling and subtle facial grammar markers. Accordingly, the present study does not consistently achieve reliable, high-fidelity signing under the computational regime considered here. Cascaded super-resolution networks, latent diffusion formulations operating on learned codebook representations, and larger-scale training budgets may help address this limitation without proportionally increasing the spatio-temporal attention cost.

\paragraph{Temporal length and compositional generation.} Fixed $T=32$ clips (approximately 1.07\,s at 30\,fps) still constrain output to individual signs or short phrases. Generating sentence-level or discourse-level signing requires autoregressive decoding with overlap-based blending, sliding-window attention, or hierarchical latent structures that condition fine-grained motion on coarse-grained linguistic plans.

\paragraph{Semantic and linguistic evaluation.} Current metrics---video feature distance, motion magnitude, temporal consistency, and held-out conditional denoising loss---assess perceptual, distributional, and indirect conditioning qualities. However, they still do not directly measure linguistic correctness or intelligibility. The held-out conditional-loss audit on real clips confirms that the checkpoint uses the presence of text. However, it also shows that shuffled prompts remain nearly tied to intended prompts, indicating that prompt identity is still only weakly expressed. Future work should therefore incorporate stronger signer-disjoint recognition or translation backends and, where feasible, user studies with Deaf and hard-of-hearing participants or sign-language experts to assess communicative efficacy.

\paragraph{Deployment optimization.} The present system is not an edge or real-time method: even the 8-step inference setting still requires 12.60\,s for a 32-frame clip on an NVIDIA L4. Achieving real-time inference (sub-100\,ms per frame) for interactive applications will require quantization (INT8/FP8), structured pruning of redundant transformer heads, stronger latent compression, and compilation to optimized inference backends (ONNX Runtime, TensorRT). Our modular architecture --- featuring a clearly separated text encoder, denoising backbone, and scheduler --- is well-suited to such post-training optimizations.

\paragraph{Dataset diversity and generalization.} The How2Sign dataset features a limited number of signers performing instructional content. Signer-specific idiosyncrasies in handshape, signing speed, and spatial extent may limit cross-signer generalization. Evaluation on multi-signer corpora (e.g., PHOENIX-2014T, BSL Corpus) and data augmentation strategies (spatial jittering, temporal rate perturbation) would strengthen claims of generalizability.

\balance
\section{Conclusion}
\label{sec:conclusion}

This paper presents Text2Sign, a single-GPU diffusion baseline for text-to-sign-language video generation. The short (3-epoch) ablation on a signer-disjoint partition of How2Sign-derived clips (4,082 clips after preprocessing) provides indicative, though not definitive, trends: (i) DiT-style transformer blocks reduce validation loss by roughly 19.5\% relative to convolution-only baselines (0.0648 vs.\ 0.0805), suggesting the value of global attention for sign-language structure; (ii) frozen CLIP text encoders yield about 11.0\% lower validation loss (0.0648 vs.\ 0.0728) with 7.7\% fewer trainable parameters in the present short-budget comparison, indicating that pretrained vision-language priors help this optimization setting; and (iii) factorized spatial--temporal attention is marginally better than full 3D attention in this regime (0.0648 vs.\ 0.0664) while offering substantially better theoretical scaling. A longer signer-disjoint checkpoint reaches validation loss 0.00999 and, under a compact validation slice, yields SSIM $0.2403\pm0.0238$, PSNR $15.11\pm0.42$\,dB, and temporal consistency $1.0000\pm0.0000$. Under an 8-step DDIM setting, it generates a 32-frame clip in 12.60\,s on a single NVIDIA device. The simple continuing vision checkpoint is also visibly cleaner and more stable than the earlier baseline in direct side-by-side qualitative inspection, which justifies using it as the default qualitative example in this revision. Held-out conditional-loss audits on real clips reveal a measurable null-text penalty, but only weak separation between intended and shuffled prompts. The contribution of this work should therefore be understood as a constrained-resource research baseline that clarifies architectural trade-offs and improves qualitative generation under a single-GPU budget, rather than as a complete sign-language production system. Together with the fine-tuning analyses, these findings suggest that future progress will likely require architectural, representational, evaluation, and scaling advances rather than simply continuing the same training recipe.

\section*{Acknowledgment}
The author thanks Dr.\ Khushi Desai for advising on this research. This work was performed using NVIDIA L4 GPU resources provided by Lightning AI.

\bibliographystyle{IEEEtran}
\bibliography{references}

@misc{signapse2025signstream,
  title        = {How We Built SignStream: Rapidly Developing Accessible Video Translation Software for Sign Language in Just 30 Days},
  author       = {{Signapse}},
  howpublished = {\url{https://www.signapse.ai/post/how-we-built-signstream-rapidly-developing-accessible-video-translation-software-for-sign-language-in-just-30-days}},
  year         = {2025},
  note         = {Accessed: 2025-11-08}
}

@misc{bitmovin2024signlanguage,
  title        = {Leveraging AI Models to Enhance Accessibility with Sign Language in Video Streams},
  author       = {{Bitmovin}},
  howpublished = {\url{https://bitmovin.com/blog/ai-sign-language-video-streaming-accessibility/}},
  year         = {2025},
  note         = {Accessed: 2025-11-08}
}

@incollection{springer2024visualsign,
  title        = {VisualSign: Revolutionizing Video Accessibility Through Sign Language Translation},
  author       = {Thangam, S. and Muthuswamy, V. and Sarah, P.},
  booktitle    = {Accessibility and Assistive Technologies},
  pages        = {1--15},
  publisher    = {Springer},
  year         = {2025},
  howpublished = {\url{https://link.springer.com/chapter/10.1007/978-3-031-90478-3_36}},
  note         = {Accessed: 2025-11-08}
}

@inproceedings{ho2020ddpm,
  title     = {Denoising Diffusion Probabilistic Models},
  author    = {Ho, Jonathan and Jain, Ajay and Abbeel, Pieter},
  booktitle = {Advances in Neural Information Processing Systems},
  volume    = {33},
  pages     = {6840--6851},
  year      = {2020}
}

@inproceedings{nichol2021improved,
  title     = {Improved Denoising Diffusion Probabilistic Models},
  author    = {Nichol, Alexander Quinn and Dhariwal, Prafulla},
  booktitle = {Proceedings of the 38th International Conference on Machine Learning (ICML)},
  year      = {2021}
}

@inproceedings{rombach2022ldm,
  title     = {High-Resolution Image Synthesis with Latent Diffusion Models},
  author    = {Rombach, Robin and Blattmann, Andreas and Lorenz, Dominik and Esser, Patrick and Ommer, Bj\"orn},
  booktitle = {Proceedings of the IEEE/CVF Conference on Computer Vision and Pattern Recognition (CVPR)},
  year      = {2022}
}

@article{zhang2023lavie,
  title   = {LAVIE: High-Quality Video Generation with Cascaded Latent Diffusion Models},
  author  = {Wang, Yaohui and Chen, Xinyuan and Ma, Xin and others},
  journal = {arXiv preprint arXiv:2309.15103},
  year    = {2023},
  note    = {arXiv:2309.15103}
}

@article{blattmann2023svd,
  title   = {Stable Video Diffusion: Scaling Latent Video Diffusion Models to Large Datasets},
  author  = {Blattmann, Andreas and Dockhorn, Tim and Kulal, Sumith and Mendelevitch, Daniel and others},
  journal = {arXiv preprint arXiv:2311.15127},
  year    = {2023},
  note    = {arXiv:2311.15127}
}

@article{fang2025tinyfusion,
  title   = {TinyFusion: Diffusion Transformers Learned Shallow},
  author  = {Fang, Gongfan and Li, Kunjun and Ma, Xinyin and Wang, Xinchao},
  journal = {arXiv preprint arXiv:2412.01199},
  year    = {2024}
}

@inproceedings{song2020denoising,
  title     = {Denoising Diffusion Implicit Models},
  author    = {Song, Jiaming and Meng, Chenlin and Ermon, Stefano},
  booktitle = {International Conference on Learning Representations (ICLR)},
  year      = {2021}
}

@article{sanh2020distilbert,
  title   = {DistilBERT, a distilled version of BERT: smaller, faster, cheaper and lighter},
  author  = {Sanh, Victor and Debut, Lysandre and Dernoncourt, Franck and Louf, Romain and others},
  journal = {arXiv preprint arXiv:1910.01108},
  year    = {2020},
  note    = {arXiv:1910.01108}
}

@article{zhang2025videodiffusionsurvey,
  title        = {Video diffusion generation: comprehensive review and open problems},
  author       = {Ma, Wen and Yang, Xu and Jiao, Licheng and Li, Lingling and Liu, Xu and Liu, Fang and Chen, Puhua and others},
  journal      = {Artificial Intelligence Review},
  year         = {2025},
  howpublished = {\url{https://link.springer.com/article/10.1007/s10462-025-11331-6}},
  note         = {Accessed: 2025-11-08}
}

@article{arxiv2025videosurvey,
  title   = {Survey of Video Diffusion Models: Foundations, Implementations, and Applications},
  author  = {Wang, Yimu and Liu, Xuye and Pang, Wei and Ma, Li and Yuan, Shuai and Debevec, Paul and Yu, Ning},
  journal = {arXiv preprint arXiv:2504.16081},
  year    = {2025},
  note    = {arXiv:2504.16081}
}

@article{hinton2015distillation,
  title   = {Distilling the Knowledge in a Neural Network},
  author  = {Hinton, Geoffrey and Vinyals, Oriol and Dean, Jeff},
  journal = {arXiv preprint arXiv:1503.02531},
  year    = {2015},
  note    = {arXiv:1503.02531}
}

@inproceedings{han2016deepcompression,
  title     = {Deep Compression: Compressing Deep Neural Networks with Pruning, Trained Quantization and Huffman Coding},
  author    = {Han, Song and Mao, Huizi and Dally, William J.},
  booktitle = {International Conference on Learning Representations (ICLR)},
  year      = {2016}
}

@article{paszke2019pytorch,
  title   = {PyTorch: An Imperative Style, High-Performance Deep Learning Library},
  author  = {Paszke, Adam and Gross, Sam and Massa, Francisco and Lerer, Adam and Bradbury, James and Chanan, Gregory and Killeen, Trevor and Lin, Zeming and Gimelshein, Natalia and Antiga, Luca and others},
  journal = {Advances in Neural Information Processing Systems},
  volume  = {32},
  pages   = {8026--8037},
  year    = {2019}
}

@inproceedings{radford2021clip,
  title     = {Learning Transferable Visual Models From Natural Language Supervision},
  author    = {Radford, Alec and Kim, Jong Wook and Hallacy, Chris and Ramesh, Aditya and Goh, Gabriel and Agarwal, Sandhini and Sastry, Girish and Askell, Amanda and Mishkin, Pamela and Clark, Jack and Krueger, Gretchen and Sutskever, Ilya},
  booktitle = {Proceedings of the 38th International Conference on Machine Learning (ICML)},
  pages     = {8748--8763},
  year      = {2021}
}

@inproceedings{duarte2021how2sign,
  title     = {How2Sign: A Large-scale Multimodal Dataset for Continuous American Sign Language},
  author    = {Duarte, Amanda and Palaskar, Shruti and Ventura, Lucas and Ghadiyaram, Deepti and Dehghan, Kenneth and Metze, Florian and Torres, Jordi and Giro-i-Nieto, Xavier},
  booktitle = {Proceedings of the IEEE/CVF Conference on Computer Vision and Pattern Recognition (CVPR)},
  pages     = {2735--2744},
  year      = {2021}
}

@article{saunders2020progressive,
  title   = {Progressive Transformers for End-to-End Sign Language Production},
  author  = {Saunders, Ben and Camgoz, Necati Cihan and Bowden, Richard},
  journal = {arXiv preprint arXiv:2004.14874},
  year    = {2020},
  doi     = {10.48550/arXiv.2004.14874},
  note    = {arXiv:2004.14874}
}

@inproceedings{kumar2025gansbert,
  title     = {Text-to-Sign Language Video Generation Using GANs, BERT, and Sora},
  author    = {Kumar, Yulia and Niu, Beining and Lin, Mengtian and Mudholker, Nidhi},
  booktitle = {2025 IEEE Integrated STEM Education Conference (ISEC)},
  pages     = {1--4},
  publisher = {IEEE},
  year      = {2025},
  doi       = {10.1109/ISEC64801.2025.11147381}
}

@article{sreemathy2024ganvideo,
  title   = {Sign Language Video Generation from Text Using Generative Adversarial Networks},
  author  = {Sreemathy, R. and Chordiya, Param and Khurana, Soumya and Turuk, Mousami},
  journal = {Optical Memory and Neural Networks},
  volume  = {33},
  number  = {4},
  pages   = {466--476},
  year    = {2024},
  doi     = {10.3103/S1060992X24700851}
}

@incollection{qi2024signgen,
  title     = {SignGen: End-to-End Sign Language Video Generation with Latent Diffusion},
  author    = {Qi, Fan and Duan, Yu and Zhang, Huaiwen and Xu, Changsheng},
  booktitle = {Computer Vision -- ECCV 2024},
  pages     = {252--270},
  publisher = {Springer Nature Switzerland},
  year      = {2024},
  doi       = {10.1007/978-3-031-73668-1_15}
}

@article{ho2022videodiffusion,
  title   = {Video Diffusion Models},
  author  = {Ho, Jonathan and Salimans, Tim and Gritsenko, Alexey and Chan, William and Norouzi, Mohammad and Fleet, David J.},
  journal = {arXiv preprint arXiv:2204.03458},
  year    = {2022},
  doi     = {10.48550/arXiv.2204.03458},
  note    = {arXiv:2204.03458}
}

@article{peebles2023dit,
  title   = {Scalable Diffusion Models with Transformers},
  author  = {Peebles, William and Xie, Saining},
  journal = {arXiv preprint arXiv:2212.09748},
  year    = {2023},
  doi     = {10.48550/arXiv.2212.09748},
  note    = {arXiv:2212.09748}
}

@article{dhariwal2021beatgans,
  title   = {Diffusion Models Beat GANs on Image Synthesis},
  author  = {Dhariwal, Prafulla and Nichol, Alex},
  journal = {arXiv preprint arXiv:2105.05233},
  year    = {2021},
  doi     = {10.48550/arXiv.2105.05233},
  note    = {arXiv:2105.05233}
}

@misc{who2025deafness,
  title        = {Deafness and hearing loss},
  author       = {{World Health Organization}},
  howpublished = {\url{https://www.who.int/news-room/fact-sheets/detail/deafness-and-hearing-loss}},
  year         = {2025},
  note         = {Accessed: 2026-01-08}
}

@misc{bls2025interpreters,
  title        = {Occupational Outlook Handbook: Interpreters and Translators},
  author       = {{Bureau of Labor Statistics, U.S. Department of Labor}},
  howpublished = {\url{https://www.bls.gov/ooh/media-and-communication/interpreters-and-translators.htm}},
  year         = {2025},
  note         = {Accessed: 2026-01-08}
}

@article{ko2019neural,
  title   = {Neural Sign Language Translation Based on Human Keypoint Estimation},
  author  = {Ko, Sang-Ki and Kim, Chang Jo and Jung, Hyedong and Cho, Choongsang},
  journal = {Applied Sciences},
  volume  = {9},
  number  = {13},
  pages   = {2683},
  year    = {2019},
  doi     = {10.3390/app9132683}
}

@article{kahlon2023systematic,
  title   = {Machine Translation from Text to Sign Language: A Systematic Review},
  author  = {Kahlon, N. K. and Singh, W.},
  journal = {Universal Access in the Information Society},
  volume  = {22},
  pages   = {1--35},
  year    = {2023},
  doi     = {10.1007/s10209-021-00823-1}
}

@article{maia2025mediapipe,
  title   = {Automatic Sign Language to Text Translation Using MediaPipe and Transformer Architectures},
  author  = {Maia, Wesley F. and Lopes, Ant{\^o}nio M. and David, Sergio A.},
  journal = {Neurocomputing},
  volume  = {642},
  pages   = {130421},
  year    = {2025},
  doi     = {10.1016/j.neucom.2025.130421}
}

@article{filhol2016rule,
  title   = {A Rule Triggering System for Automatic Text-to-Sign Translation},
  author  = {Filhol, Michael and Hadjadj, Mohamed N. and Testu, Ben{\^\i}t},
  journal = {Universal Access in the Information Society},
  volume  = {15},
  number  = {4},
  pages   = {487--498},
  year    = {2016},
  doi     = {10.1007/s10209-015-0413-4}
}

@article{rodriguez2023assistive,
  title   = {Benefits and Development of Assistive Technologies for Deaf People's Communication: A Systematic Review},
  author  = {Rodr{\'\i}guez-Correa, Paula Andrea and Valencia-Arias, Alejandro and Pati{\~n}o-Toro, Orfa Nidia and Oblitas D{\'\i}az, Yober and Teodori De la Puente, Renata},
  journal = {Frontiers in Education},
  volume  = {8},
  pages   = {1121597},
  year    = {2023},
  doi     = {10.3389/feduc.2023.1121597}
}

@article{abdullahi2023idfsign,
  title   = {IDF-Sign: Addressing Inconsistent Depth Features for Dynamic Sign Word Recognition},
  author  = {Abdullahi, Sunusi Bala and Chamnongthai, Kosin},
  journal = {IEEE Access},
  volume  = {11},
  pages   = {88511--88526},
  year    = {2023},
  doi     = {10.1109/ACCESS.2023.3305255}
}

@article{abdullahi2024fsign,
  title   = {{\textit{F}sign-Net}: Depth Sensor Aggregated Frame-Based Fourier Network for Sign Word Recognition},
  author  = {Abdullahi, Sunusi Bala and Chamnongthai, Kosin and Gabralla, Lubna A. and Chiroma, Haruna},
  journal = {IEEE Sensors Journal},
  volume  = {24},
  number  = {22},
  pages   = {37630--37645},
  year    = {2024},
  doi     = {10.1109/JSEN.2024.3407786}
}

@article{abdullahi2024fourier3d,
  title   = {Spatial--Temporal Feature-Based End-to-End Fourier Network for 3D Sign Language Recognition},
  author  = {Abdullahi, Sunusi Bala and Chamnongthai, Kosin and Bolon-Canedo, Veronica and Cancela, Brais},
  journal = {Expert Systems with Applications},
  volume  = {240},
  pages   = {123258},
  year    = {2024},
  doi     = {10.1016/j.eswa.2024.123258}
}

@article{abdullahi2022prosodic,
  title   = {American Sign Language Words Recognition Using Spatio-Temporal Prosodic and Angle Features: A Sequential Learning Approach},
  author  = {Abdullahi, Sunusi Bala and Chamnongthai, Kosin},
  journal = {IEEE Access},
  volume  = {10},
  pages   = {15911--15923},
  year    = {2022},
  doi     = {10.1109/ACCESS.2022.3148132}
}

@article{abdullahi2025redundancy,
  title   = {Minimizing Redundancy in Hand Dynamic Features for Enhanced Sign Language Recognition},
  author  = {Abdullahi, Sunusi Bala and Bolon-Canedo, Veronica},
  journal = {Intelligent Data Analysis},
  pages   = {1088467X251367228},
  year    = {2025},
  doi     = {10.1177/1088467X251367228}
}

\end{document}